\documentclass[journal]{IEEEtran}
\usepackage{graphicx}
\usepackage{algorithm} 
\usepackage{algorithmic} 
\usepackage{color}
\usepackage{amsmath}
\usepackage{CJK}
\usepackage{flushend}
\usepackage{amsfonts}
\usepackage{mathrsfs}
\usepackage{booktabs}
\usepackage{hyperref}
\usepackage{multirow}
\usepackage{cite}
\ifCLASSINFOpdf
\else
\fi

\begin{document}
\title{Ensemble Super-Resolution with A Reference Dataset}

\author{Junjun~Jiang,~\IEEEmembership{Member,~IEEE,}
        Yi~Yu,
        Zheng Wang,
        Suhua~Tang,~\IEEEmembership{Member,~IEEE,}
        Ruimin Hu,~\IEEEmembership{Senior Member,~IEEE}
        and Jiayi Ma,~\IEEEmembership{Member,~IEEE}

\IEEEcompsocitemizethanks{
\IEEEcompsocthanksitem J. Jiang is with the School of Computer Science and Technology, Harbin Institute of Technology, Harbin 150001, China, and is also with the Peng Cheng Laboratory, Shenzhen, China (jiangjunjun@hit.edu.cn).
\IEEEcompsocthanksitem Y. Yu and Z. Wang are with the Digital Content and Media Sciences Research Division, National Institute of Informatics, Tokyo 101-8430, Japan (\{yiyu, wangz\}@nii.ac.jp).
\IEEEcompsocthanksitem S. Tang is with the Department of Communication Engineering and Informatics, The University of Electro-Communications, Tokyo 182-8585, Japan (shtang@uec.ac.jp).
\IEEEcompsocthanksitem R. Hu is with the National Engineering Research Center for Multimedia Software, School of Computer, Wuhan University, Wuhan, 430072, China (hrm1964@163.com).
\IEEEcompsocthanksitem J. Ma is with the Electronic Information School, Wuhan University, Wuhan 430072, China (jyma2010@gmail.com).
}

}




\maketitle

\begin{abstract}
By developing sophisticated image priors or designing {deep(er) architectures}, a variety of image Super-Resolution (SR) approaches have been proposed recently and achieved very promising performance. A natural question that arises is whether these methods can be reformulated into a unifying framework and whether this framework assists in SR reconstruction? In this paper, we present a simple but effective single image SR method based on ensemble learning, which can produce a better performance than that could be obtained from any of SR methods to be ensembled (or called \emph{component super-resolvers}). Based on the assumption that better component super-resolver should have larger ensemble weight when performing SR reconstruction, we present a Maximum A Posteriori (MAP) estimation framework for the inference of optimal ensemble weights. Specially, we introduce a reference dataset, which is composed of High-Resolution (HR) and Low-Resolution (LR) image pairs, to measure the super-resolution abilities (prior knowledge) of different component super-resolvers. To obtain the optimal ensemble weights, we propose to incorporate the reconstruction constraint, which states that the degenerated HR image should be equal to the LR observation one, as well as the prior knowledge of ensemble weights into the MAP estimation framework. Moreover, the proposed optimization problem can be solved by an analytical solution. We study the performance of the proposed method by comparing with different competitive approaches, including four state-of-the-art non-deep learning based methods, four latest deep learning based methods and one ensemble learning based method, and {prove} its effectiveness and superiority on {three} public datasets.
\end{abstract}


\IEEEpeerreviewmaketitle

\section{Introduction}

Image Super-Resolution (SR) is a class of image processing technology which can infer a High-Resolution (HR) image from one or a sequence of Low-Resolution (LR) images~\cite{Park2013SPM}. It can transcend the limitations of current optical imaging systems, and has been widely applied in medical and remote sensing imaging, digital photographs, depth based 3D reconstruction, and intelligent video surveillance system~\cite{Wang2013Survey,liu2019depthSR,jiangK2018deep}.

The SR problem is a severely ill-posed inverse problem due to information loss during the image degradation process, \emph{e.g.}, image blurring, aliasing from subsampling and noise.
How to reconstruct an HR image which looks pleasant from an LR one remains an extremely challenging task. The prior knowledge, such as piecewise smoothness~\cite{aly2005image, Liu_TIP14_denoising, Liu2017Random}, shape edges~\cite{sun2008image, Chen2017HXH}, textures~\cite{hacohen2010image}, local/nonlocal similar patterns~\cite{Glasner2009, Liu2011Image, zhang2011interpolation, jiang2017TMMsingle}, low-rank constraint~\cite{DongWS2013TIP, GgongWG2015INS}, and sparse representations under certain transformations~\cite{Yang2010TIP, yang2012compressive, LiXY2015TIP, yang2016curvelet, ma2013regularized}, have been investigated to regularize the SR reconstruction procedures. Generally speaking, the current methods fall into two general categories: multi-frame reconstruction approaches and learning-based single image SR approaches.

By making full use of the inter-frame complementary information, multi-frame reconstruction based SR approaches leverage a sequence of LR images of the same scene and fuses them to induce an HR output or a sequence of HR outputs. However, the sub-pixel registration is an exceedingly difficult problem and the magnification factor is limited in practice~\cite{Lin2004}. Learning-based single image SR methods aim at learning the relationship between the LR and HR example pairs, and then {applying} the learned transformation to predict missing details of an observed LR image. In this paper, we focus on the single image SR problem.

Since the pioneer work by Freeman \emph{et al.}~\cite{Freeman2002}, single image SR problem has increasingly been studied and attracted great research interests in recent decades. For example, Chang \emph{et al.}~\cite{Chang2004NE} introduced the locally linear embedding~\cite{Roweis2000} based manifold learning theory into SR problem for the first time, and then a series of neighbor embedding algorithms have been proposed~\cite{sun2008image, Bevilacqua2012, Yang2014dual, ma2015robust,ma2018nonrigid}. They can well exploit the local manifold structure of image patch space. To adaptively select the neighbor samples, Yang \emph{et al.}~\cite{Yang2010TIP} proposed to use {sparse} representation algorithm to adaptively choose the most relevant neighbors, avoiding over- or under-fitting of these neighbor embedding based method and obtaining better results~\cite{Zhang2011close, Kim2010PAMI, Zeyde2012, Jiang2015PJ}. 
In order to overcome the inconsistency between the LR and HR spaces, quite a few coupled learning based methods have also been developed recently~\cite{JiaWT13, Wang2012Semi, YangWM2016TMM}. They are essentially in order to learn the relationship from one domain/space to another domain/space, \emph{i.e.}, from the LR space to the corresponding HR one. The approach of Timofte \emph{et al.} \cite{Timofte2013ICCV} leverages the \emph{divide and conquer} strategy to learn the mapping relationship between the LR and HR samples in multiple local neighbor spaces, and a fast single image SR method based on Anchored Neighborhood Regression (ANR) is developed. In order to further enhance the quality of mapping relationship, they further combine ANR with simple function based method~\cite{yang2013fast} and proposed the Adjusted ANR (A+ for short) approach~\cite{timofte2014a}. A+ studies the mapping relationship between the LR and HR samples in a much denser sample space, which can guarantee the performance of local linear regression. In addition to the work of~\cite{Timofte2013ICCV, yang2013fast, timofte2014a}, some regression algorithms also have been developed to directly learn the relationship between the LR samples and HR samples in a coarse-to-fine~\cite{Zhang2016Coarse, WangNN2016TIP}, sparse~\cite{Tang2014Learning, Yang2017Single, LiuJY2017TMM}, collaborative~\cite{ZhangCCR2016TMM, Zhang2017Collaborative}, adaptive~\cite{Chen2017HXH}, local~\cite{Yang2017Joint, ShenHF2016Adaptive}, pairwise~\cite{ShaoLing2017TIP} or structured~\cite{Deng2016Similarity} manner. The above mentioned algorithms are simple, fast, and can well characterize the potential mapping between the LR and HR spaces (especially the local image patch space), and thus they produced very favorable performance.

Over the past few years, deep learning, the re-emergence of neural networks, has been tremendously and successfully used in a multitude of fields, such as self-driving cars, computer vision, speech recognition, and machine translation, and has achieved significant and impressive results~\cite{lecun2015deep}. Most recently, this technology has also been introduced to solve the image SR problem by learning the mapping relationship between the LR and HR samples in an \emph{end-to-end} manner~\cite{dong2016image, huang2015single, kim2016accurate, kim2016deeply, Zeng2017CoupledDeep, Huang2017DOTE,wang2015deep,WangZY2019TIP}.  Super-Resolution using Deep Convolutional Networks (SRCNN)~\cite{dong2016image}, Cascade of Sparse Coding based Networks (CSCN)~\cite{wang2015deep}, Very Deep Convolutional Networks (VDSR)~\cite{kim2016accurate}, and Deeply-Recursive Convolutional Networks (DRCN)~\cite{kim2016deeply} based deep learning SR techniques carefully design different network structures to meet the challenge of SR reconstruction. Specifically, SRCNN~\cite{dong2016image} constructs a three convolutional layers, while CSCN~\cite{wang2015deep} cascades sparse coding networks. In~\cite{kim2016accurate}, VDSR makes use of the deep model up to 20 weights layers to predict residual image between the HR images and LR ones. By this very deep network, it can use large receptive field and take a large image context into account, thus well capturing the image structure especially when the scale factor increase. DRCN~\cite{kim2016deeply} recursively leverages the same convolutional network as many times as desired while does not introduce additional parameters for additional convolutions. To get better human perception, a number of photo-realism based Generative Adversarial Networks (GAN)~\cite{goodfellow2014generative} have also been presented newly~\cite{johnson2016perceptual, Yuan_2018_CVPR_Workshops}. 

However, the aforementioned methods based on different shallow prior models (local manifold structure prior or sparse prior) or different deep networks have their own advantages and capture different image details.
Over the years, we have witnessed a constant effort to design a better performance for the SR problem. A natural question that arises is \emph{whether these methods can be reformulated into a unifying framework and whether this framework assists in SR task?}

One very natural idea is to integrate the outputs of different SR methods (we call the SR algorithms to be ensembled as \emph{component super-resolvers} in the following) in an ensemble learning framework and produce an output that is better than all component super-resolvers. Then, given a number of results obtained by the component super-resolvers, how to ensemble them to produce a better result? The most obvious way is directly averaging all the component super-resolvers equally. However, ensemble learning theory~\cite{zhou2002ensembling} has proved that \emph{it may be better to combine some instead of all of the learners}.
That is to say, when we know in advance that the performance of one component super-resolver is poor, we can remove it or set a relative small ensemble weigh in advance. So, the remaining question is how to determine whether a component super-resolver is superior or not. In other words, how to determine the ensemble weights is the essential problem in ensemble learning based SR problem. 

\begin{figure}
  \centering
  \includegraphics[width=8.8cm]{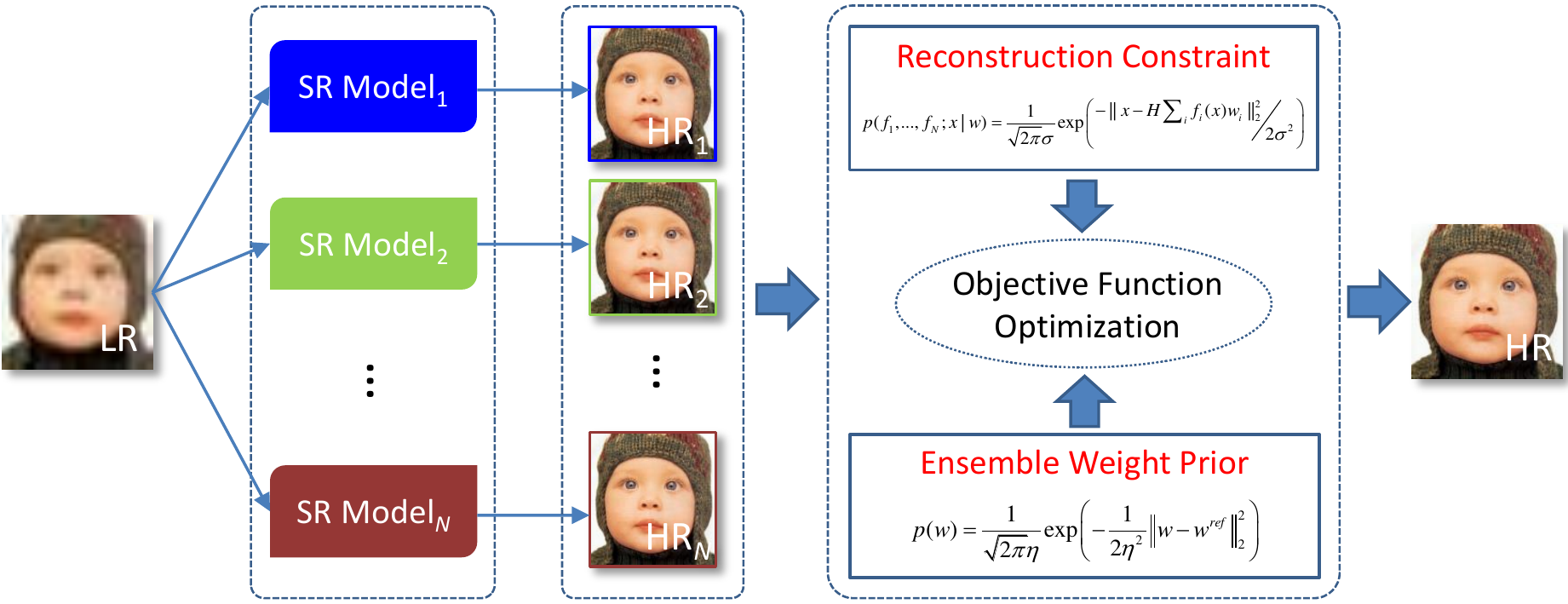}\\
  \caption{The basic idea of the proposed RefESR method. Given an LR observation image, we generate its various HR version by different SR models. To produce the ensemble HR output, we solve the ensemble weights by an analytical approach, meeting the reconstruction constraint and the ensemble weight prior learned from an additional reference dataset.}\label{img:framework}
\end{figure}

In this paper, we contribute a simple but effective \textbf{E}nsemble learning \textbf{SR} algorithm with a \textbf{Ref}erence dataset, which is denoted as \emph{RefESR} for short. Our method is inspired by external dataset based models. Unlike previously methods that learn prior knowledge for the parameters of one statistical model or the desired HR images, our method directly learn the SR abilities of different methods and use them to guide the optimization of ensemble parameters, \emph{i.e.}, the ensemble (or combination) weights. To estimate the optimal ensemble weights, in particular, the proposed RefESR method considers both the posterior reconstruction error deduced from the image degradation model and the ensemble weight prior learned from an additional reference dataset, and formulates them in a Maximum A Posteriori (MAP) framework. Moreover, we introduce a simple method to obtain an analytical solution of the ensemble parameters. Fig. \ref{img:framework} shows the pipeline of the proposed RefESR algorithm. To the best of our knowledge, this is the first time to leverage an additional reference dataset to guide the SR reconstruction. Although many previous works have presented to use an additional dataset to exploit the natural image prior, our proposed method directly leverages a reference dataset to obtain the {SR ability (in terms of objective qualities)} of different component super-resolvers, and applies it to guide the subsequence SR reconstruction. Experimental results demonstrate that our RefESR method is better than state-of-the-art deep learning based SR methods. Moreover, our method is very general and it can be used to ensemble the best methods fed into our framework to improve the SR performance, thus expecting to always achieve the best reconstruction results.

The following paragraphs of this paper are organized as follows: In Section \ref{sec:related}, we present some related works of ensemble learning based SR approaches. Section \ref{sec:proposed} introduces the proposed ensemble SR framework and the objective function optimization method in detail. The experimental results are presented in Section \ref{sec:experiments}. Some deep analysis and discussions to the proposed ensemble learning framework are presented in Section \ref{sec:dis}. Finally, we conclude this work in Section \ref{sec:conclusion}.

\section{Related Work}
\label{sec:related}
In statistics and machine learning, ensemble learning method is a powerful way to produce a better performance than that could be obtained from any of the component methods. It has been widely applied in the fields of data mining and pattern recognition~\cite{zhou2012ensemble}. Although ensemble learning has achieved great success in machine learning problems, it has not been applied to image SR. Until most recently, two ensemble learning related SR methods have been proposed.

In \cite{liao2015video}, a video SR method is presented. They decompose the video SR task into two stages: draft-ensemble generation and determine the optimal one via convolutional neural network deep learning. In essence, they leveraged the deep learning networks to select the candidate HR samples in the patch space, and this is the general idea of lots of learning-based SR methods. Through it is termed as ensemble-based, it is not strictly ensemble SR method because selecting the best samples for the following reconstruction is the basic idea of many learning based SR methods~\cite{Freeman2002, Chang2004NE, Yang2010TIP}. The other work is proposed by Wang \emph{et al.}~\cite{wang2017ensemble}, they introduced the ensemble learning into the SR problem and proposed an ensemble based deep networks method for image SR. It focuses on one deep learning based SR method, and generates different models by different initializations of one specific neural network. Specifically, they took sparse coding based networks~\cite{wang2015deep} as baseline, and developed an Ensemble based Sparse Coding Networks (ESCN) by changing the initializations of SCN~\cite{wang2015deep}. In ESCN, the ensemble weights are adaptively determined by a back-projection model.

ESCN based SR method has achieved better performance than the original SCN method~\cite{wang2015deep}, however, there are two limitations: Firstly, it essentially integrates only one deep learning model, SCN based neural network, with multiple outputs under different initial conditions. Unfortunately, due to the limited capacity of the same network, the complementary information obtained by only changing the initialization is insufficient, thus the improvement of the ensemble result is limited. Secondly, it only considers the reconstruction constraints when determining the ensemble weights and no other prior has been taken into consideration. Their model is actually ill-posed, and there are many solutions to meet its objective function. From their experiments we can also find that the optimal ensemble weights and average weights obtained almost the same results. Therefore, it is not really effective to consider only reconstruction constraints. In contrast, our proposed method ensembles a variety of  different methods, {including traditional state-of-the-art learning based methods} and deep learning based methods with different neural networks emerged in recent years. Moreover, we introduce a reference dataset to measure the performance of different SR methods, which can be seen as the model prior and is incorporated into to our objective function as a regularization term.

\section{Proposed Method}
\label{sec:proposed}
In this section, we present the proposed RefESR method in detail. We firstly give the problem definition of RefESR in a Bayesian framework. Then, we show how to model the reconstruction constraint and the prior of ensemble weights. And then, we induce out the objective function of our proposed RefESR method. After that, we describe an analytical way to solve the optimization problem.

\subsection{Problem Setup}
In our proposed ensemble learning based SR method, we can obtain the SR reconstruction results, $\textbf{y}_1,\textbf{y}_2,...,\textbf{y}_N$, of different methods, $f_1,f_2,...,f_N$, for the observed LR image, $\textbf{x}$. Here, ${f_i}(x) \to {y_i}$ can be seen as the $i$-th SR model. Given $f_1,f_2,...,f_N$ and $\textbf{x}$ in the ensemble SR framework, our aim is to infer the optimal ensemble weights, ${\textbf{w}^*} = {[w_1^*,w_2^*, \cdots ,w_N^*]^T}$, where $w_i^*$ is associated with the $i$-th SR model $f_i$. After obtaining the optimal ensemble weights, we can predict the HR output of LR input $\textbf{x}_t$ by
\begin{equation}\label{eq:prediction}
{\textbf{y}_t} = \sum\limits_{i = 1}^N {{f_i}({\textbf{x}_t})w_i^*}.
\end{equation}
Under the Bayesian framework, the regularized SR problem is related to a probabilistic model as follows:
\begin{equation}\label{eq:map}
\begin{array}{c}
\begin{split}
 p(\textbf{w}|{f_1},{f_2}, \cdots ,{f_N};\textbf{x}) = \frac{{p({f_1},{f_2}, \cdots ,{f_N};\textbf{x}|\textbf{w})p(\textbf{w})}}{{p({f_1},{f_2}, \cdots ,{f_N};\textbf{x})}} \\
\propto p({f_1},{f_2}, \cdots ,{f_N};\textbf{x}|\textbf{w})p(\textbf{w}).
\end{split}\nonumber
 \end{array}
\end{equation}
Notice that the marginal likelihood, ${p({f_1},{f_2}, \cdots ,{f_N};\textbf{x})}$, does not depend on $\textbf{w}$. With the observation of $f_1,f_2,...,f_N$ and $\textbf{x}_t$, the MAP estimation of $\textbf{w}$ can be formulated as,
\begin{equation}\label{eq:map2}
\begin{split}
 {\textbf{w}^*}& = \mathop {\arg \max }\limits_\textbf{w} p(\textbf{w}|{f_1},{f_2}, \cdots ,{f_N};\textbf{x}) \\
&= \mathop {\arg \max }\limits_\textbf{w} \left\{ {p({f_1},{f_2}, \cdots ,{f_N};\textbf{x}|\textbf{w})p(\textbf{w})} \right\} \\
&= \mathop {\arg {\rm{min}}}\limits_\textbf{w} \left\{ { - {\rm{log }}p({f_1},{f_2}, \cdots ,{f_N};\textbf{x}|\textbf{w}) - {\rm{log}}{\kern 1pt} p(\textbf{w})} \right\}, \\
 \end{split}
\end{equation}
where the first term is the likelihood term and the second term denotes the prior knowledge of the  $\textbf{w}$. By the definition of the likelihood term and the prior term, we can {maximize} the objective function (\ref{eq:map}) to obtain the optimal ensemble weights $\textbf{w}^{*}$. Acquiring the optimal ensemble weights, we can expect to infer the target HR output.

\subsection{Reconstruction Constraint Modeling}
For single image SR problem, the relationship between the HR image \textbf{y} and the LR one \textbf{x} can be modeled by the observation model~\cite{elad1997restoration}:
\begin{equation}\label{eq:degenaration}
\textbf{x = DBy + v}.	
\end{equation}
Here, we denote the matrix $\textbf{B}$ a blurring operator, the matrix $\textbf{D}$ a matrix representing the down-sampling operator, and the matrix $\textbf{v}$ the additive Gaussian white noise. If we use the matrix $\textbf{H}$ to denote the blurring and {downsampling} processes ({the matrix $\textbf{H}$ stands for the degradation operations}), (\ref{eq:degenaration}) can be rewritten as~\cite{Park2013SPM}:
\begin{equation}\label{eq:degenaration2}
\textbf{x = Hy + v}.
\end{equation}

Since the matrix $\textbf{H}$ has far fewer rows than columns, Eq. (\ref{eq:degenaration2}) is ill-posed and has an infinite number of solutions. Therefore, in order to recover a reasonable HR image, SR approaches typically try to find and model an appropriate prior knowledge of natural images. For example, gradient prior, self-similarity property (that some salient features repeat across different scales within an image), or the coupled LR/HR patches based algorithms have been used to effectively model the prior for building the inverse recovery mapping problem.

Developing sophisticated image priors has been the focus of much single image SR research in the past decade. In contrast, the reconstruction constraint, which states that the degenerated HR image should be equal to the LR observation one, has received relatively little attention. Some algorithms do not enforce $\textbf{x = Hy}$ at all. The representative ANR~\cite{Timofte2013ICCV}, A+~\cite{timofte2014a}, and recently proposed deep learning based methods~\cite{dong2016image, huang2015single, kim2016accurate, kim2016deeply} all ignore this reconstruction constraint.

To this end, in our ensemble learning based SR framework, we introduce this reconstruction constraint to our objective function. Specially, we enforce the blurred and downsampled HR ensemble output should approximately equal the low-res input image. We assume that the difference between the ensemble HR output and the LR input image, \emph{i.e.}, the reconstruction error obeys the Gaussian distribution, thus the likelihood term can be written as follows
\begin{equation}\label{eq:Constraint2}
\small
p({f_1},{f_2}, \!\cdots \! ,{f_N};\textbf{x}|\textbf{w}) \!=\! \frac{1}{{\sqrt {2\pi } \sigma }}\exp \left( { - \frac{1}{{2{\sigma ^2}}}\left\| {\textbf{x} \!-\! H\sum\limits_{i = 1}^N {{f_i}(\textbf{x}){w_i}} } \right\|_2^2} \right),{\kern 1pt} {\kern 1pt}
\end{equation}
where $\sigma$ denotes the standard deviation of the noise.

\begin{figure}
  \centering
  \includegraphics[width=8.8cm]{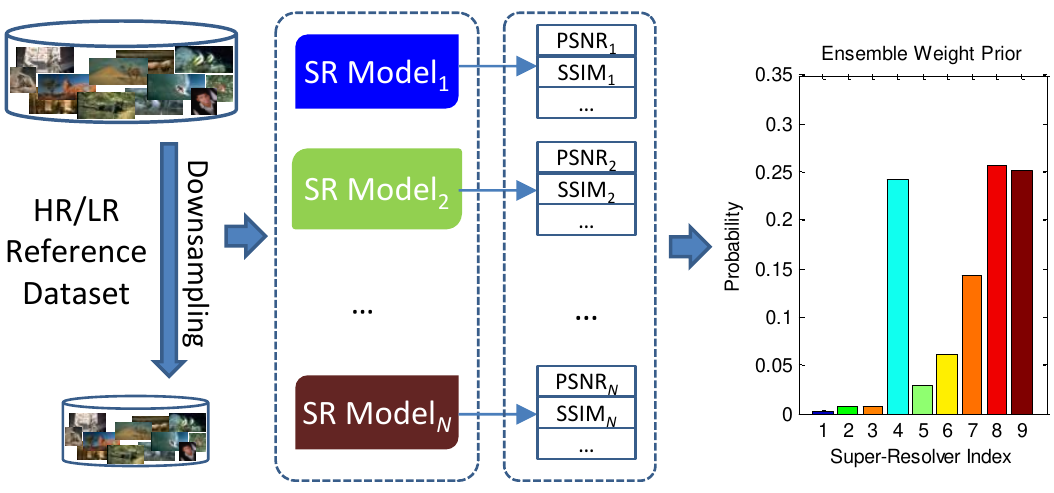}\\
  \vspace{-0.20cm}
  \caption{Learning the prior of ensemble weights. We first generate the reference dataset by collecting a large number of HR images, and downsampling them to obtain the corresponding LR images. For each LR reference image, we predict its HR version by each component super-resolver, \emph{i.e.}, different SR models. And then, we apply the objective evaluation metrics to the predicted HR image and the ground truth to calculate the scores of each super-resolver. Lastly, the prior of ensemble weights can be obtained according to Eq. (\ref{eq:wref0}).}\label{img:WeightsPrior}
\end{figure}

\subsection{Prior Modeling of Ensemble Weights}
The aforementioned reconstruction constraint can be seen as the specific regularization for the ensemble weights of an observed LR image. In this subsection, we propose to regularize the ensemble weights by defining another prior of the ensemble weights, thus overcoming the ill-posed solution of Eq. (\ref{eq:Constraint2}).

In practice, the performance of component super-resolver is unknown. However, we can get their SR results on a reference dataset, which can be used to approximate the performance. Specifically, we introduce an additional reference dataset, and then test the performance of component super-resolvers. Then, their reconstruction quality evaluations $scor{e_i}$ can be obtained by combining their performances at different magnification factors, \emph{e.g.}, 2, 3, and 4 in our experiments,
\begin{equation}\label{eq:score}
scor{e_i} = \sum\limits_{s = 2}^4 {psnr(i,s) \cdot ssim(i,s)}, {\kern 12pt}  i=1,2,\cdots, N.\nonumber
\end{equation}
We denote $psnr(i,s)$ and $ssim(i,s)$ the mean {Peak Signal-to-Noise Ratio (PSNR) and Structural Similarity Index (SSIM)~\cite{Wang2004SSIM} results of the $i$-th component super-resolver at scale $s$, respectively. It is worth mentioning that more measurements can be incorporated to obtain the performance score. Our basic assumption is that the method obtaining a better performance on the reference dataset should get a relatively larger weight when reconstructing the HR output image of an LR input one in the ensemble framework. Fig.~\ref{img:WeightsPrior} shows the process of obtaining the ensemble weight prior.

\begin{figure}
  \centering
  \includegraphics[width=0.49\textwidth]{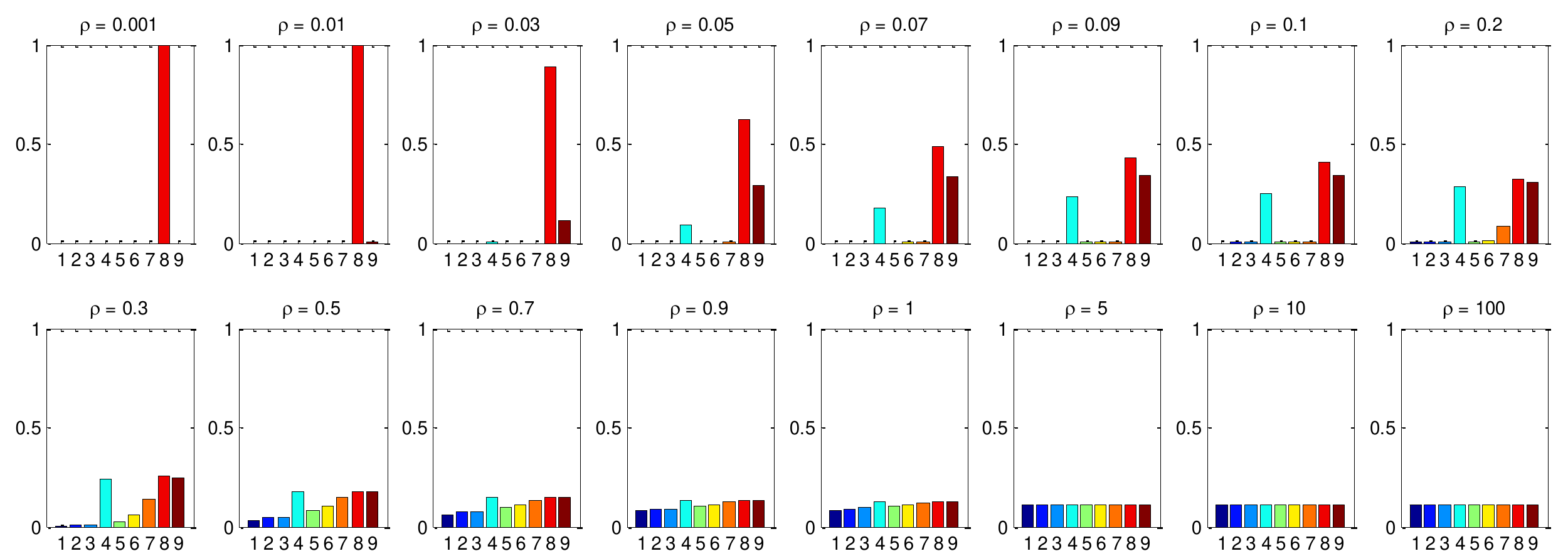}\\
  \vspace{-0.20cm}
  \caption{The influences of the bandwidth parameter $\rho $ on the reference ensemble weights. The abscissa is index of the component super-resolver, while the ordinate is the reference ensemble weight. Index 1 to 9 are the methods of Kim~\cite{Kim2010PAMI}, SelfExSR~\cite{huang2015single}, A+~\cite{timofte2014a}, and IA~\cite{timofte2016seven}, SRCNN~\cite{dong2016image}, CSCN~\cite{wang2015deep}, CSCN-MV~\cite{wang2015deep}, VDSR~\cite{kim2016accurate}, and DRCN~\cite{kim2016deeply}.}
  \label{img:weightsRho}
\end{figure}

Therefore, given the performance of component super-resolvers on the reference dataset, we define the $i$-th element of the reference weight vector ${\textbf{w}^{{\rm{r}}ef}} = {[w_1^{{\rm{re}}f},w_2^{{\rm{re}}f}, \cdots ,w_N^{{\rm{re}}f}]^T}$ as follows,
\begin{equation}\label{eq:wref0}
w_i^{{\rm{re}}f} = \frac{{\exp \left( { - {\raise0.7ex\hbox{${{{\left( {scor{e_i} - scor{e_{\max }}} \right)}^2}}$} \!\mathord{\left/
 {\vphantom {{{{\left( {scor{e_i} - scor{e_{\max }}} \right)}^2}} {\rho^2}}}\right.\kern-\nulldelimiterspace}
\!\lower0.7ex\hbox{${\rho^2}$}}} \right)}}{{\sum\nolimits_{i = 1}^N {\exp \left( { - {\raise0.7ex\hbox{${{{\left( {scor{e_i} - scor{e_{\max }}} \right)}^2}}$} \!\mathord{\left/
 {\vphantom {{{{\left( {scor{e_i} - scor{e_{\max }}} \right)}^2}} {\rho^2}}}\right.\kern-\nulldelimiterspace}
\!\lower0.7ex\hbox{${\rho ^2}$}}} \right)} }},
\end{equation}
where $\rho $ is the bandwidth parameter, and $scor{e_{\max }}$ is the best performance of $N$ component super-resolvers, $scor{e_{\max }} = \max \{ scor{e_i},scor{e_2}, \cdots ,scor{e_N}\}$. The numerator represents the performance similarity between $i$-th component super-resolver and the best method, while the denominator is normalization constant used to guarantee the sum of all element of $\textbf{w}^{ref}$ to be one. The bandwidth parameter $\rho $ is crucial for the following SR task. {Very large or small values} will be detrimental to the final result. As shown in Fig. \ref{img:weightsRho}, when the value of $\rho $ is too small, the best component super-resolver will dominate the SR reconstruction, \emph{i.e.}, the weight of the best component super-resolver will be close to 1, while other component super-resolvers are almost 0. In contrast, when the value of $\rho $ is too large, all the component super-resolvers will contribute equally to the SR reconstruction, \emph{i.e.}, different component super-resolvers are assigned to the same weights. For more detailed analysis, please refer to the experimental section.

Note that $\textbf{w}^{ref}$ denotes the prior weights learned from the reference dataset. Our aim is to obtain an input specific ensemble weight vector $\textbf{w}$ that cannot differ too much from $\textbf{w}^{ref}$. Thus, we can define the prior probability of $\textbf{w}$ by Gaussian model due to its simplicity:
\begin{equation}\label{eq:Prior}
p(\textbf{w}) = \frac{1}{{\sqrt {2\pi } \eta }}\exp \left( { - \frac{1}{{2{\eta ^2}}}\left\| {\textbf{w} - {\textbf{w}^{ref}}} \right\|_2^2} \right),
\end{equation}
where $\eta$ is a scale parameter for the prior distribution of ensemble weights $\textbf{w}$.

\subsection{Objective Function}
By substituting Eq. (\ref{eq:Constraint2}) and Eq. (\ref{eq:Prior}) into the Eq. (\ref{eq:map2}) and dropping some constant terms, we have
\begin{equation}\label{eq:Function1}
\begin{array}{c}
 {\textbf{w}^*} \! =  \!\mathop {\arg \min\limits_\textbf{w}  - \log \frac{1}{{\sqrt {2\pi } \sigma }}\exp \left( { - \frac{1}{{2{\sigma ^2}}}\left\| {\textbf{x}  \!-  \!\textbf{H}\sum\limits_{i = 1}^N {{f_i}(\textbf{x}){w_i}} } \right\|_2^2} \right)}\limits_\textbf{w}  \\
  - \log \frac{1}{{\sqrt {2\pi } \eta }}\exp \left( { - \frac{1}{{2{\eta ^2}}}\left\| {\textbf{w} - {\textbf{w}^{ref}}} \right\|_2^2} \right) \\
  = \mathop {\arg \min }\limits_\textbf{w} \left\| {\textbf{x} - \textbf{H}\sum\limits_{i = 1}^N {{f_i}(\textbf{x}){w_i}} } \right\|_2^2 + \lambda \left\| {\textbf{w} - {\textbf{w}^{ref}}} \right\|_2^2.\\
  \end{array}
\end{equation}
The first term is the reconstruction error, while the second is the difference between a pre-learned weight vector ${\textbf{w}^{ref}}$ and the optimal weight vector to be estimated. The regularization parameter $\lambda$ is related to $\sigma$ and $\eta$ by $\lambda  = \frac{{{\sigma ^2}}}{{{\eta ^2}}}$, and is used to balance the contributions between the reconstruction error and the prior knowledge of $\textbf{w}$.

In order to make the ensemble SR results interpretable, we present to incorporate the sum-to-one constraint to the objective function. Thus, we have
\begin{equation}\label{eq:Function2}
\begin{array}{c}
 {\textbf{w}^*} = \mathop {\arg \min }\limits_\textbf{w}  {\left\| {\textbf{x} - \textbf{H}\sum\limits_{i = 1}^N {{f_i}(\textbf{x}){w_i}} } \right\|_2^2 + \lambda \left\| {\textbf{w} - {\textbf{w}^{ref}}} \right\|_2^2} , \\
 {\kern 1pt} {\kern 1pt} {\kern 1pt} {\kern 1pt} {\rm{s}}{\rm{.t}}{\rm{.}}{\kern 1pt} {\kern 1pt} {\kern 1pt} {\kern 1pt} \sum\limits_{i = 1}^N {{w_i}}  = 1.\\
 \end{array}
\end{equation}

To obtain an optimal ensemble weight vector, we simultaneously take into consideration the input dependent reconstruction constraint and the prior of the ensemble methods learned from a reference dataset. {The first term can be seen a global reconstruction constraint, which can guarantee the consistence between the degraded HR estimation and the input LR image. For these patch based SR methods~\cite{Chang2004NE,Timofte2013ICCV,timofte2014a}, the averaged and fused HR estimation may not meet perfectly with the global reconstruction constraint ~\cite{Yang2010TIP, Gao2012Joint}. In other words, these patches based SR methods reconstruct the HR image locally (patchwise) and ignore the global information. Through adding this global reconstruction constraint, our method can guarantee the degraded HR image (\textbf{Hy}) is equal to the observed LR image (\textbf{x}), and thus capturing more information about the global structure of the target HR image.} Therefore, the proposed ensemble model can avoid the problem of lack of flexibility due to the absence of data-based reconstruction constraints, or the problem of the solution is not unique due to ill-posed conditions. 

\subsection{Optimization}
For the blurring and downsampling processes are the liner operator, thus we have
\begin{equation}\label{eq:linear}
\textbf{H}\sum\limits_{i = 1}^N {{f_i}(\textbf{x}){w_i}}  = \sum\limits_{i = 1}^N {\left( {\textbf{H}{f_i}(\textbf{x})} \right){w_i}}  = {\bf{Yw}}.
\end{equation}
Each column of $\bf{Y}$ denotes one downsampled HR output, ${\bf{Y}} = [{\textbf{H}}{f_1}(\textbf{x}),{\textbf{H}}{f_2}(\textbf{x}),\cdots,{\textbf{H}}{f_N}(\textbf{x})]$. By substituting Eq. (\ref{eq:linear}) to Eq. (\ref{eq:Function2}), the objective function can be rewritten as the following matrix form,
\begin{equation}\label{eq:matrixFunction2}
\begin{array}{c}
 {\textbf{w}^*} = \mathop {\arg \min }\limits_\textbf{w}  {\left\| {\textbf{x} - \textbf{Yw} } \right\|_2^2 + \lambda \left\| {\textbf{w} - {\textbf{w}^{ref}}} \right\|_2^2} , \\
 {\kern 1pt} {\kern 1pt} {\kern 1pt} {\kern 1pt} {\rm{s}}{\rm{.t}}{\rm{.}}{\kern 1pt} {\kern 1pt} {\kern 1pt} {\kern 1pt} \sum\limits_{i = 1}^N {{w_i}}  = 1.\\
 \end{array}
\end{equation}

Eq. (\ref{eq:matrixFunction2}) can be written as,
\begin{equation}\label{eq:combined}
\begin{array}{c}
 {{\bf{w}}^*} = \mathop {\arg \min }\limits_{\bf{w}} \left\{ {\left\| {{{\bf{x}}^{'}} - {{\bf{Y}}^{'}}} {\bf{w}}\right\|_2^2} \right\},
 {\kern 1pt} {\kern 1pt} {\kern 1pt} {\kern 1pt} {\rm{s}}{\rm{.t}}{\rm{.}}{\kern 1pt} {\kern 1pt} {\kern 1pt} {\kern 1pt} \sum\limits_{i = 1}^N {{w_i}}  = 1,
 \end{array}
\end{equation}
where and $\bf{x^{'}}={\left[ \begin{array}{l}
 {\kern 1pt} {\kern 1pt} {\kern 1pt} {\kern 1pt} {\kern 1pt} {\kern 1pt} {\kern 1pt} {\kern 1pt} {\kern 1pt} {\kern 1pt} {\kern 1pt} {\bf{x}} \\
 \sqrt \lambda  {{\bf{w}}^{ref}} \\
 \end{array} \right]}$, $\bf{Y^{'}}={\left[ \begin{array}{l}
 {\kern 1pt} {\kern 1pt} {\kern 1pt} {\kern 1pt} {\kern 1pt} {\bf{Y}} \\
 \sqrt \lambda  {\bf{I}} \\
 \end{array} \right]}$, and $\bf{I}$ is a unit matrix with the size of $N\times N$.

Apparently, Eq. (\ref{eq:combined}) is a constrained linear least squares problem. {Following the work of \cite{Roweis2000}}, we first define a local Gram matrix $\textbf{G}$ for ${\bf{x}}^{'}$,
\begin{equation}\label{eq:Gram}
\textbf{G} = ({\bf{x}}^{'}\textbf{1}^T-\bf{Y^{'}})^T({\bf{x}}^{'}\textbf{1}^T-\bf{Y^{'}}),
\end{equation}
where $\textbf{1}$ is a $N\times 1$ column vector of ones. Then, the problem (\ref{eq:combined}) has the following analytical solution:
\begin{equation}\label{eq:solutin}
{\bf{w}^{*}} = \frac{{{{\bf{G}}^{ - 1}}{\bf{1}}}}{{{{\bf{1}}^T}{{\bf{G}}^{ - 1}}{\bf{1}}}}.
\end{equation}
Upon acquiring the optimal ensemble weights of $\textbf{w}^{*}$, we can just simply combine the results of component super-resolvers $\{f_i(\textbf{x})\}_{i=1}^N$ and $\textbf{w}^{*}$ through Eq.~(\ref{eq:prediction}). {It is worth noting that the objective functions of Cevikalp \emph{et al}. \cite{Cevikalp2008} and our proposed method are essentially a constrained least squares problem, as proposed in \cite{Roweis2000}. The work of \cite{Cevikalp2008} tries to obtain the optimal combination weights of different classifiers to achieve the best classification performance, while our method focuses on the image SR problem, and tries to obtain the optimal combination (ensemble) weights with the global reconstruction constraint as well as the prior of the weight constraint. In this sense, they are different though they all use the same optimization method to solve their objective function. In fact, in the field of image processing and computer vision, the objective function of many methods is a very simple, \emph{i.e.}, a constrained least squares problem. The difference lies in that different methods use different constraints (prior knowledge) to regularize the solutions. How to find a good prior knowledge and how to model it effectively is the key to the success of an algorithm. The novelty of the proposed method is the introduction of a reference dataset and using it to produce prior knowledge to regularize the combination (ensemble) weights.}

\begin{table}[t]
\centering
\caption{PSNR (dB) and SSIM comparisons with state-of-the-arts on \textbf{SET14}. The best and second best results are marked in \textcolor{red}{red} and \textcolor{blue}{blue}, respectively.}
\label{tab:PSNRSSIMSet14}
\vspace{1pt}
\scriptsize
\begin{tabular}{|c||c|c|c|c|c|c|}
\hline
Dataset      & \multicolumn{6}{c|}{\textbf{SET14}}	\\
\hline
Scale &	\multicolumn{2}{c|}{$\times2$} &	\multicolumn{2}{c|}{$\times 3$} & \multicolumn{2}{c|}{$\times 4$}         	\\
\hline
Metric &	PSNR&SSIM&PSNR&SSIM&PSNR&SSIM\\
\hline
\hline
Bicubic	&	30.24	&	0.8688	&	27.55	&	0.7742	&	26.00	&	0.7027		\\
Kim	\cite{Kim2010PAMI}&	32.14	&	0.9032	&	28.96	&	0.8144	&	27.18&	0.744			\\
SelfExSR	\cite{huang2015single}&	32.22	&	0.9034	&	29.16	&	0.8196	&	27.40	&	0.7518	\\
A+	\cite{timofte2014a}&	32.28	&	0.8056	&	29.13	&	0.8188	&	27.32	&	0.7491			\\
IA	\cite{timofte2016seven}&	32.83	&	0.9110	&	29.63	&	0.8296	&	27.85	&	0.7643		\\
SRCNN	\cite{dong2016image}&	32.42	&	0.9063	&	29.28	&	0.8209	&	27.49	&	0.7503	 \\
CSCN	\cite{wang2015deep}&	32.56	&	0.9074	&	29.41	&	0.8238	&	27.64	&	0.7587		\\
CSCN-MV	\cite{wang2015deep}&	32.80	&	0.9101	&	29.57	&	0.8263	&	27.81	&	0.7619		\\
VDSR	\cite{kim2016accurate}&	33.03	&	\textcolor[rgb]{0.00,0.00,1.00}{0.9124}	&	\textcolor[rgb]{0.00,0.00,1.00}{29.78}	&	\textcolor[rgb]{0.00,0.00,1.00}{0.8314}	&	28.01	&	\textcolor[rgb]{0.00,0.00,1.00}{0.7674}\\
DRCN	\cite{kim2016deeply}&	\textcolor[rgb]{0.00,0.00,1.00}{33.04}	&	0.9118	&	29.77	&	0.8312	&	\textcolor[rgb]{0.00,0.00,1.00}{28.02}	&	0.7570	\\
ESCN	\cite{wang2017ensemble}&	32.67	&	0.9093	&	29.51	&	0.8264	&	27.75	&	0.7611	\\
RefESR	&	\textcolor[rgb]{1.00,0.00,0.00}{33.16}	&	\textcolor[rgb]{1.00,0.00,0.00}{0.9134}	&	\textcolor[rgb]{1.00,0.00,0.00}{29.90}	&	\textcolor[rgb]{1.00,0.00,0.00}{0.8338}	&	\textcolor[rgb]{1.00,0.00,0.00}{28.14}	&	\textcolor[rgb]{1.00,0.00,0.00}{0.7702}	\\
\hline
\end{tabular}
\end{table}

\begin{table}[t]
\centering
\caption{PSNR (dB) and SSIM comparisons with state-of-the-arts on the \textbf{SET5}. The best and second best results are marked in \textcolor{red}{red} and \textcolor{blue}{blue}, respectively.}
\label{tab:PSNRSSIMSet5}
\vspace{1pt}
\scriptsize
\begin{tabular}{|c||c|c|c|c|c|c|}
\hline
Dataset      & \multicolumn{6}{c|}{\textbf{SET5}}\\
\hline
Scale &        	\multicolumn{2}{c|}{$\times 2$} & \multicolumn{2}{c|}{$\times 3$} &	\multicolumn{2}{c|}{$\times 4$} \\
\hline
Metric &	PSNR&SSIM&PSNR&SSIM&PSNR&SSIM\\
\hline
\hline
Bicubic	&		33.66	&	0.9299	&	30.39	&	0.8682	&	28.42	&	0.8104	\\
Kim	\cite{Kim2010PAMI}&		36.24	&	0.9518	&	32.3	&	0.9032	&	30.07	&	0.8542	\\
SelfExSR	\cite{huang2015single}&		36.49	&	0.9537	&	32.58	&	0.9093	&	30.31	&	0.8619	\\
A+	\cite{timofte2014a}&		36.54	&	0.9544	&	32.59	&	0.9088	&	30.28	&	0.8603	\\
IA	\cite{timofte2016seven}&		37.37	&	0.9582	&	33.43	&	0.9186	&	31.05	&	0.8764	\\
SRCNN	\cite{dong2016image}&		36.66	&	0.9542	&	32.58	&	0.9093	&	30.86	&	0.8732	\\
CSCN	\cite{wang2015deep}&		36.93	&	0.9552	&	33.10	&	0.9144	&	30.86	&	0.8732	\\
CSCN-MV	\cite{wang2015deep}&		37.21	&	0.9571	&	33.34	&	0.9173	&	31.14	&	0.8189	\\
VDSR	\cite{kim2016accurate}&		37.53	&	0.9587	&	33.66	&	0.9213	&	31.35	&	0.8838	\\
DRCN	\cite{kim2016deeply}&		\textcolor[rgb]{0.00,0.00,1.00}{37.63}	&	\textcolor[rgb]{0.00,0.00,1.00}{0.9588}	&	\textcolor[rgb]{0.00,0.00,1.00}{33.82}	&	\textcolor[rgb]{1.00,0.00,0.00}{0.9226}	&	\textcolor[rgb]{0.00,0.00,1.00}{31.53}	&	\textcolor[rgb]{1.00,0.00,0.00}{0.8854}	\\
ESCN	\cite{wang2017ensemble}&		37.14	&	0.9571	&	33.28	&	0.9173	&	31.02	&	0.8774	\\
RefESR	&	\textcolor[rgb]{1.00,0.00,0.00}{37.71}	&	\textcolor[rgb]{1.00,0.00,0.00}{0.9593}	&	\textcolor[rgb]{1.00,0.00,0.00}{33.87}	&	\textcolor[rgb]{0.00,0.00,1.00}{0.9224}	&	\textcolor[rgb]{1.00,0.00,0.00}{31.55}	&	\textcolor[rgb]{0.00,0.00,1.00}{0.8848}	\\
\hline
\end{tabular}
\end{table}

\begin{table}[t]
\centering
\caption{PSNR (dB) and SSIM comparisons with state-of-the-arts on the \textbf{Urban100}. The best and second best results are marked in \textcolor{red}{red} and \textcolor{blue}{blue}, respectively.}
\label{tab:PSNRSSIMUrban100}
\vspace{1pt}
\scriptsize
\begin{tabular}{|c||c|c|c|c|c|c|}
\hline
Dataset      & \multicolumn{6}{c|}{\textbf{Urban100}}\\
\hline
Scale &        	\multicolumn{2}{c|}{$\times 2$} & \multicolumn{2}{c|}{$\times 3$} &	\multicolumn{2}{c|}{$\times 4$} \\
\hline
Metric &	PSNR&SSIM&PSNR&SSIM&PSNR&SSIM\\
\hline
\hline
Bicubic	&	26.88	&	0.8403	&	24.46	&	0.7349	&	23.14	&	0.6577	\\
Kim	\cite{Kim2010PAMI}&	28.71	&	0.8942	&	25.24	&	0.7761	&	23.53	&	0.6790	\\
SelfExSR	\cite{huang2015single}&	29.54	&	0.8967	&	26.44	&	0.8088	&	24.79	&	0.7374	\\
A+	\cite{timofte2014a}&	29.20	&	0.8938	&	26.03	&	0.7973	&	24.32	&	0.7183	\\
IA	\cite{timofte2016seven}&	29.93	&	0.9077	&	26.71	&	0.8106	&	24.93	&	0.7416	\\
SRCNN	\cite{dong2016image}&	29.50	&	0.8946	&	26.24	&	0.7989	&	24.52	&	0.7221	\\
CSCN	\cite{wang2015deep}&	29.14	&	0.8988	&	25.58	&	0.7858	&	23.80	&	0.6924	\\
CSCN-MV	\cite{wang2015deep}&	29.30	&	0.9015	&	25.70	&	0.7903	&	23.91	&	0.6984	\\
VDSR	\cite{kim2016accurate}&	\textcolor[rgb]{0.00,0.00,1.00}{30.76}	&	\textcolor[rgb]{0.00,0.00,1.00}{0.9140}	&	27.14	&	\textcolor[rgb]{0.00,0.00,1.00}{0.8279}	&	\textcolor[rgb]{0.00,0.00,1.00}{25.18}	&	\textcolor[rgb]{0.00,0.00,1.00}{0.7524}	\\
DRCN	\cite{kim2016deeply}&	30.75	&	0.9133	&	\textcolor[rgb]{0.00,0.00,1.00}{27.15}	&	0.8276	&	25.14	&	0.7510	\\
ESCN	\cite{wang2017ensemble}&	29.25	&	0.8986	&	25.72	&	0.7912	&	23.99	&	0.6975	\\
RefESR	&	\textcolor[rgb]{1.00,0.00,0.00}{30.88}	&	\textcolor[rgb]{1.00,0.00,0.00}{0.9150}	&	\textcolor[rgb]{1.00,0.00,0.00}{27.26}	&	\textcolor[rgb]{1.00,0.00,0.00}{0.8285}	&	\textcolor[rgb]{1.00,0.00,0.00}{25.28}	&	\textcolor[rgb]{1.00,0.00,0.00}{0.7529}	\\
\hline
\end{tabular}
\end{table}

\section{Experimental Results}
\label{sec:experiments}
In this section, we present the experimental settings used to evaluate the proposed RefESR approach and show the reconstruction results generated by carrying out SR experiments on {three} public general image databases and some face image databases. 

\subsection{Experimental Setup}
\textbf{\emph{Database}}. To test the performance, we leverage {three} commonly used image sets, \textbf{SET5}, \textbf{SET14}, and \textbf{Urban100}, as the testing images\footnote{{\textbf{SET14} includes 14 different scenes and was firstly used by Zeyde \emph{et al}.~\cite{Zeyde2012} to show their results, \textbf{SET5} includes 5 different scenes of image and was used by Bevilacqua \emph{et al}.~\cite{Bevilacqua2012}, and \textbf{Urban100} is created by Huang \emph{et al}. \cite{huang2015single} and contains 100 HR images with a variety of real-world structures, such as urban, city, and architecture. The length and width of original HR images for \textbf{SET14} (the first column), \textbf{SET5} (the second column), and \textbf{Urban100} (the last two columns) databases, are all from 200 pixels to 600 pixels.}}. Like many state-of-the-art single image SR methods~\cite{huang2015single, timofte2014a, timofte2016seven, dong2016image, wang2015deep, kim2016accurate, kim2016deeply}, in our experiments the original HR images are degenerated by Bicubic interpolation (\emph{i.e.}, the \emph{imresize} function in Matlab) with a factor of 2, 3, and 4, to generate the corresponding LR images. {It should be noted that if the image degradation process of the input LR image is unknown, which can be seen as the blind image SR, the performance of our method will reduce sharply because the mismatch between the true image degradation and simulated image degradation of the training dataset~\cite{timofte2017ntire}.} 

{Note that there are some contextual connections between the images in the reference set to image in the test set. This has been confirmed by many domain-specific image SR methods, i.e., face hallucination and text SR. When super-resolving LR face images, a good general image SR method which is trained by diversity general images is usually worse than a domain-specific face image SR method which is trained by face images. In this paper, we consider only the general image SR problem, so we hope that the reference dataset should be as diversity as possible.}

\textbf{\emph{Implementation Details}}. To ensemble different component super-resolvers, we first select some state-of-the-art SR algorithms, which include four non- deep learning, \emph{e.g.}, Kim~\cite{Kim2010PAMI}, SelfExSR~\cite{huang2015single}, A+~\cite{timofte2014a}, and IA~\cite{timofte2016seven}, and five deep learning based methods, \emph{e.g.}, SRCNN~\cite{dong2016image}, CSCN~\cite{wang2015deep}, CSCN-MV~\cite{wang2015deep}, VDSR~\cite{kim2016accurate}, and DRCN~\cite{kim2016deeply}.\footnote{We select these nine methods for their representative, pleasurable performance, and also public availability of their source codes.} Then we test their performance on the reference dataset. Because we know the ground truth of the input LR image, the SR abilities of these algorithms can be measured by some objective metrics, such as PSNR, SSIM, or their combination. And then, the reference weight vector (calculated by Eq. (\ref{eq:wref0})) is applied to regularize the optimization of the ensemble weights.

In the testing phase, we first reconstruct the HR images of above-mentioned component super-resolvers. And then, the optimal ensemble weights $\textbf{w}^*$ is obtained by Eq. (\ref{eq:solutin}). The final HR output can be constructed by the combination of the HR resultant images of different component super-resolvers and the optimal ensemble weights.

\subsection{Parameter Analysis}
In this subsection, we analyze the effect of model parameters for the performance of RefESR, and validate the proposed reconstruction constraint and reference ensemble weight prior used in the proposed network. Particularly, we conduct experiments on the testing image set of \textbf{SET14} and the magnification is 3. For other cases, we can still draw a similar conclusion. Therefore, here we will not show up one by one. From the objective function (\ref{eq:Function2}) of our method, we learn that the bandwidth parameter $\rho$ and the regularization parameter $\lambda$ have a great impact on the performance of the algorithm.

\begin{figure}
  \centering
  \includegraphics[width=7.8cm]{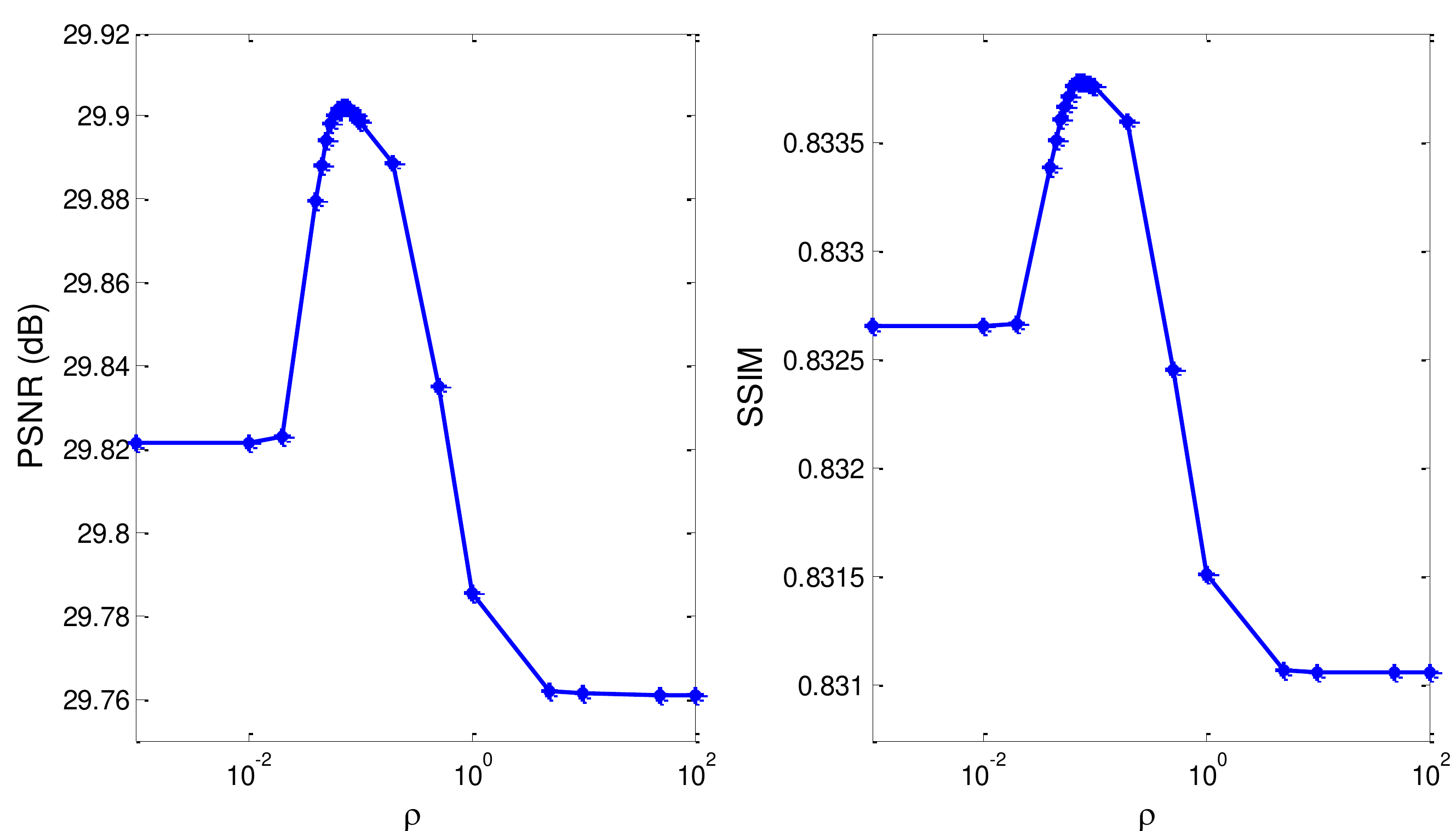}\\
  \caption{{Plots of average PSNR and SSIM results ($\times 3$ magnification) over \textbf{SET14} according to different values of the bandwidth parameter $\rho$. It achieves the best performance when $\rho$ is set to 0.07.}}
  \label{img:PlotRho}
\end{figure}

\begin{figure}
  \centering
  \includegraphics[width=7.8cm]{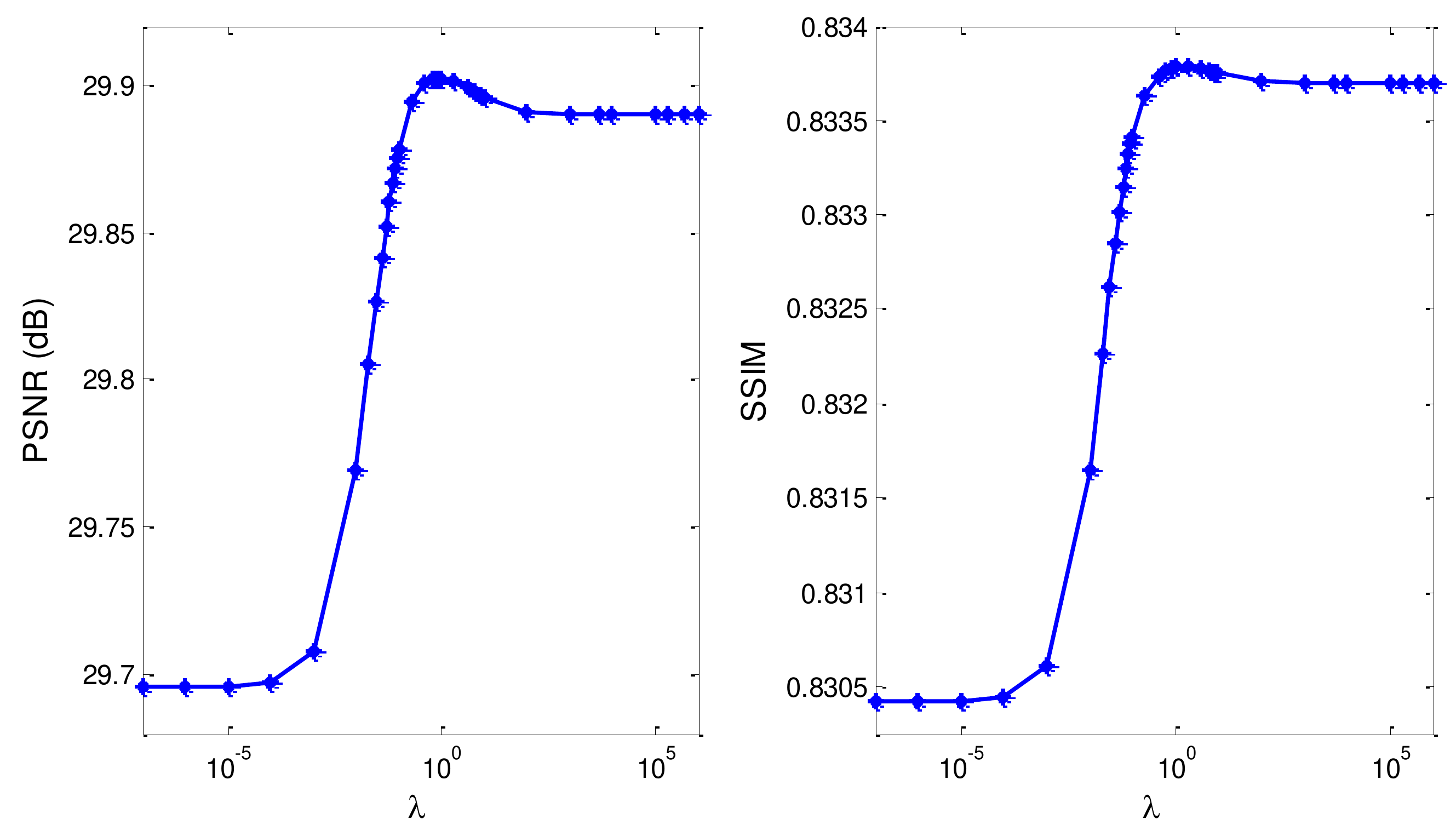}\\
  \caption{{Plots of average PSNR and SSIM results ($\times 3$ magnification) over \textbf{SET14} according to different values of the regularization parameter $\lambda$. It achieves the best performance when $\lambda$ is set to 0.8.}}
  \label{img:PlotLambda}
\end{figure}

\begin{table}[t]
\centering
 \caption{\label{tab:test}PSNR (dB) and SSIM comparisons of the proposed method with different conditions.}
 \begin{tabular}{l|ccc}
  \hline
  Method&PSNR&SSIM\\
  \hline
    \hline
Best Component Super-Resolver &29.78 &0.8314 \\
  Without Reconstruction Constraint & 29.89 & 0.8337 \\
  Without Weights Prior & 29.70 & 0.8304 \\
  Ensemble Via Averaging &29.71 &0.8301 \\
  \hline
 \bfseries{The Proposed Method} & \bfseries{29.90} & \bfseries{0.8338}\\
  \hline
 \end{tabular}
\end{table}

 \begin{figure*}
  \centering
  \includegraphics[width=0.825\textwidth]{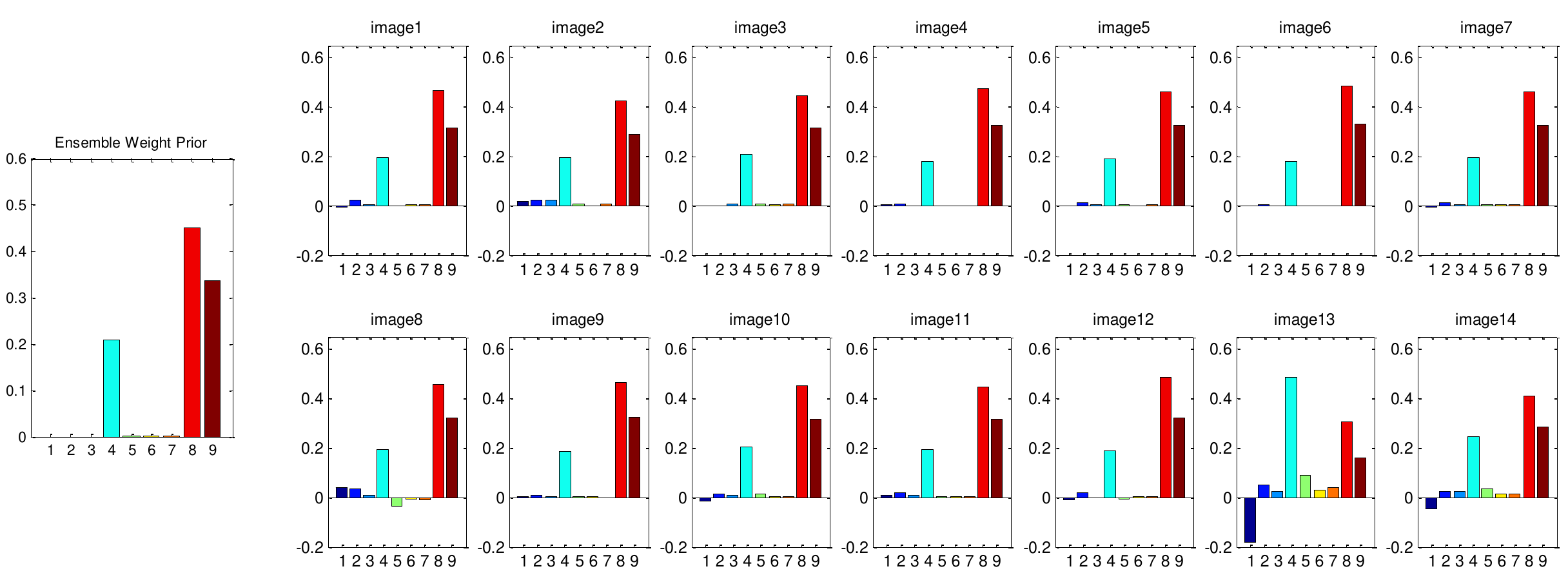}\\
  \caption{The leftmost subfigure is the initial weights obtained from the reference dataset, while the rest ones denote the optimal ensemble weights of all 14 testing images on the \textbf{SET14}. The abscissa is the same meaning as in Fig. \ref{img:weightsRho}, while the ordinate is the ensemble weight.}
  \label{img:weight14}
  \vspace{-0.20cm}
\end{figure*}

\begin{figure*}[!htpb]
\centering
\includegraphics[width=0.90\textwidth]{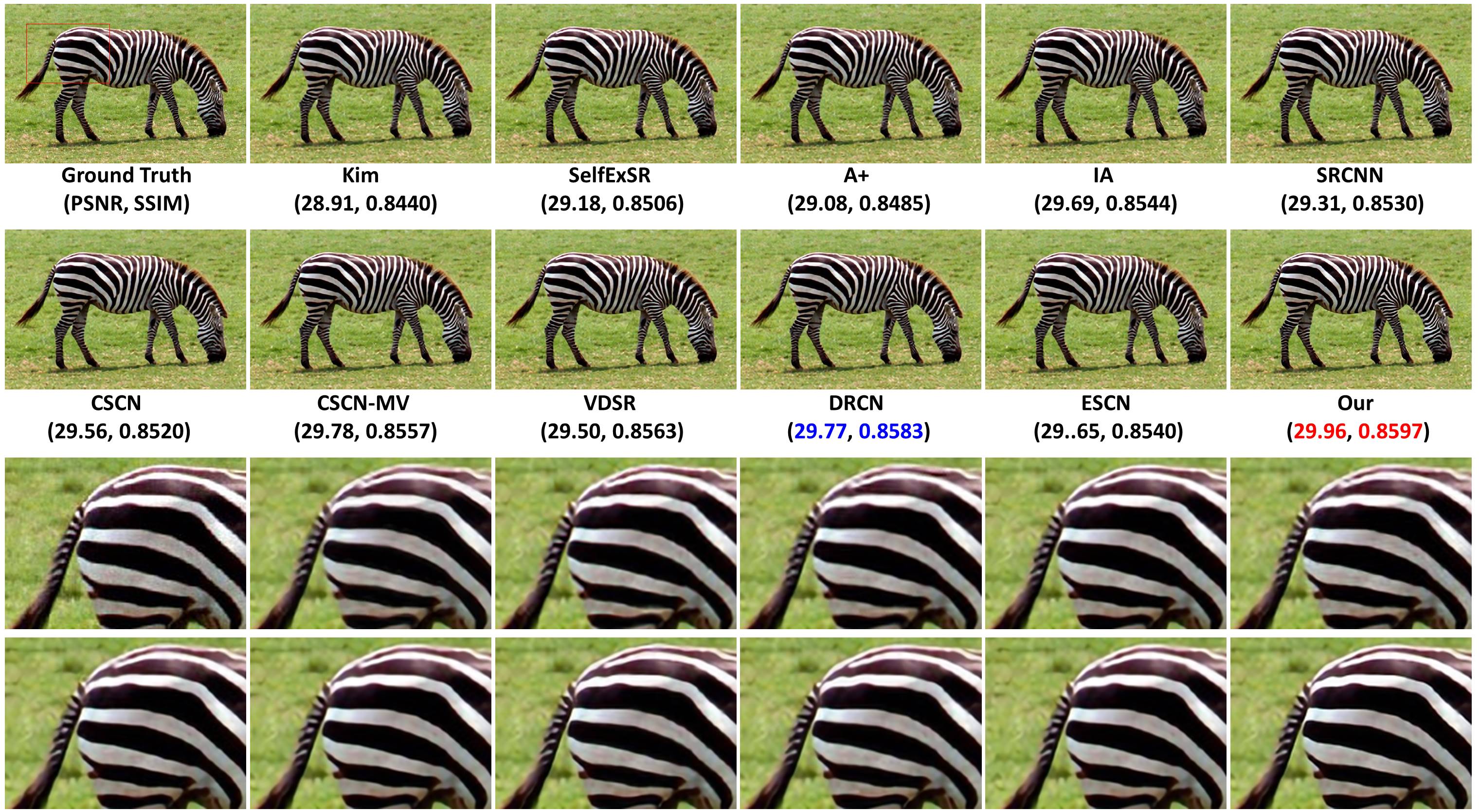}\\
\vspace{-0.20cm}
\caption{{{(Best zoomed in on screen.)} Visual reconstruction results of image ``zebra'' with a scale factor of $\times 3$. From left to right, and up to bottom: ground truth HR image, reconstructed HR image by Kim~\cite{Kim2010PAMI}, SelfExSR~\cite{huang2015single}, A+~\cite{timofte2014a}, and IA~\cite{timofte2016seven}, SRCNN~\cite{dong2016image}, CSCN~\cite{wang2015deep}, CSCN-MV~\cite{wang2015deep}, VDSR~\cite{kim2016accurate}, DRCN~\cite{kim2016deeply}, ESCN \cite{wang2017ensemble}, and our proposed RefESR method.}}
\label{fig:zebra}
\end{figure*}

\begin{figure*}[!htpb]
\centering
\includegraphics[width=0.90\textwidth]{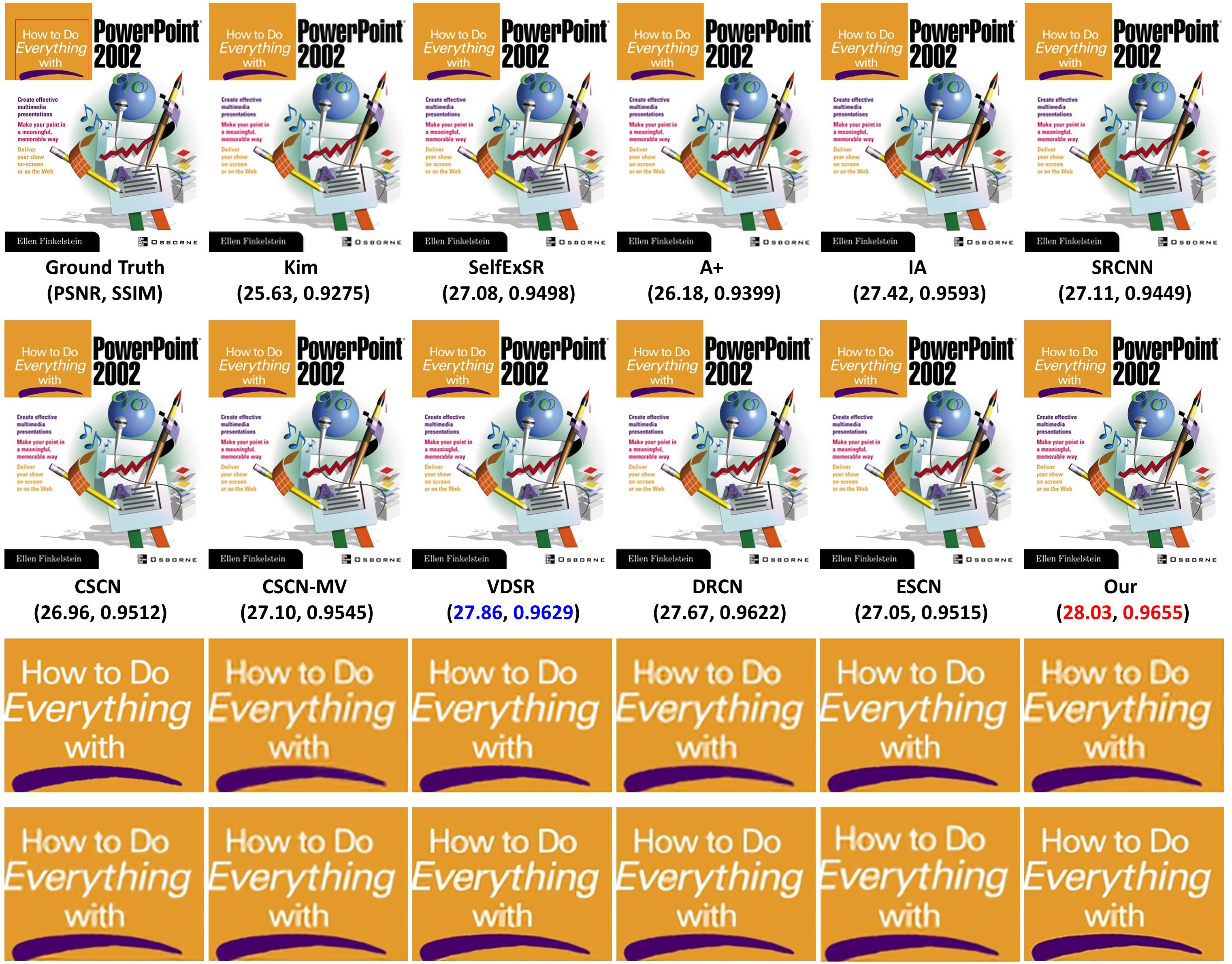}\\
\vspace{-0.20cm}
\caption{{{(Best zoomed in on screen.)} Visual reconstruction results of image ``ppt'' with a scale factor of $\times 3$. From left to right, and up to bottom: ground truth HR image, reconstructed HR image by Kim~\cite{Kim2010PAMI}, SelfExSR~\cite{huang2015single}, A+~\cite{timofte2014a}, and IA~\cite{timofte2016seven}, SRCNN~\cite{dong2016image}, CSCN~\cite{wang2015deep}, CSCN-MV~\cite{wang2015deep}, VDSR~\cite{kim2016accurate}, DRCN~\cite{kim2016deeply}, ESCN \cite{wang2017ensemble}, and our proposed RefESR method.}}
\label{fig:ppt}
\end{figure*}

\begin{figure*}[!htpb]
\centering
\includegraphics[width=0.90\textwidth]{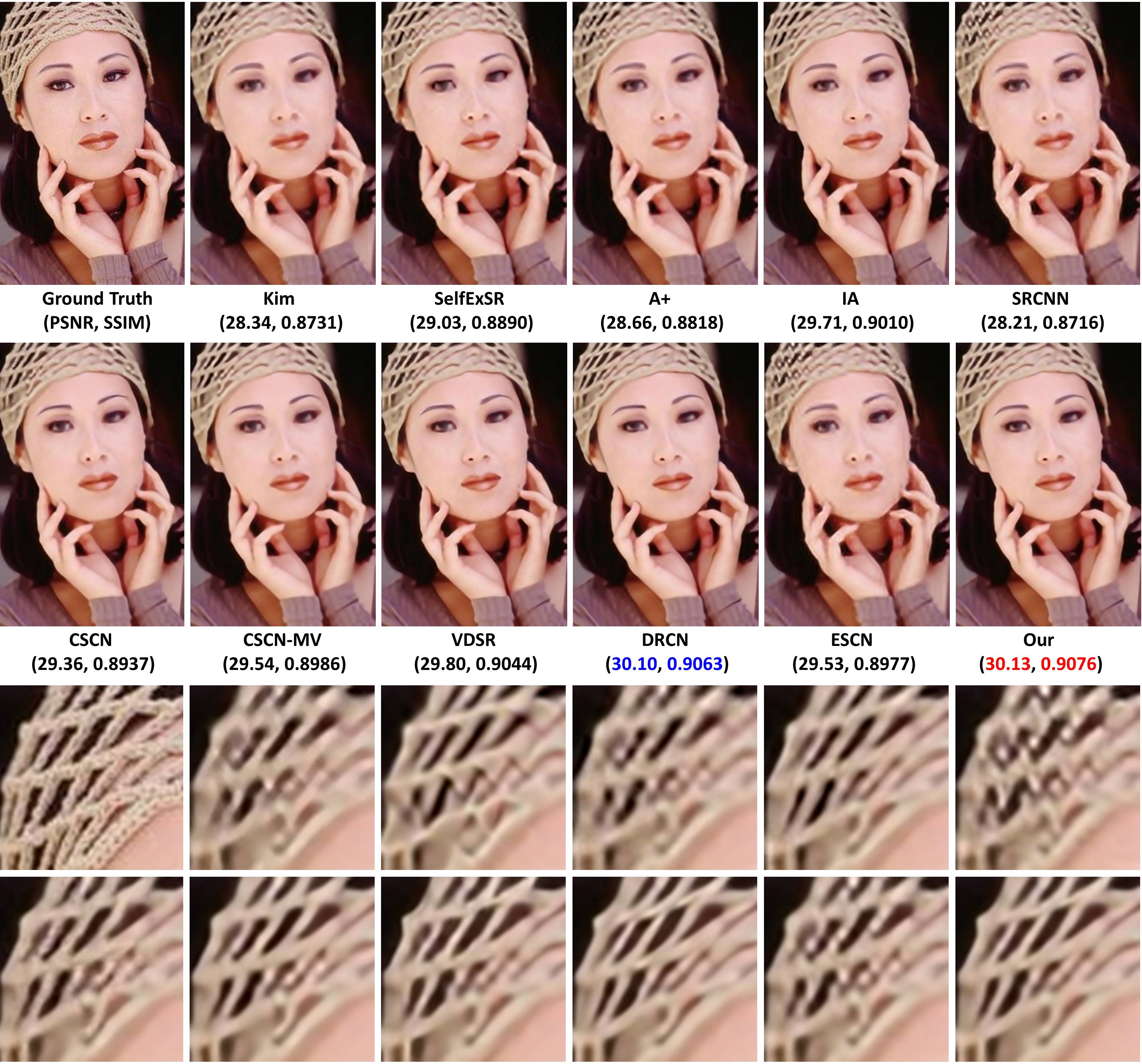}\\
\vspace{-0.20cm}
\caption{{{(Best zoomed in on screen.)} Visual reconstruction results of image ``woman'' with a scale factor of $\times 4$. From left to right, and up to bottom: ground truth HR image, reconstructed HR image by Kim~\cite{Kim2010PAMI}, SelfExSR~\cite{huang2015single}, A+~\cite{timofte2014a}, and IA~\cite{timofte2016seven}, SRCNN~\cite{dong2016image}, CSCN~\cite{wang2015deep}, CSCN-MV~\cite{wang2015deep}, VDSR~\cite{kim2016accurate}, DRCN~\cite{kim2016deeply}, ESCN \cite{wang2017ensemble}, and our proposed RefESR method.}}
\label{fig:woman}
\end{figure*}

Fig. \ref{img:PlotRho} and Fig. \ref{img:PlotLambda} show the performance of our method when the other parameter is set to the optimal. As shown in Fig. \ref{img:PlotRho}, we can at least draw the following two conclusions: (i) \emph{The ensemble SR reconstruction is effective}. This can be concluded by comparing the results when $\rho=0.001$ and $\rho=0.07$. When $\rho=0.001$, almost only the best component super-resolver is active (the 8-th method, \emph{i.e.}, VDSR~\cite{kim2016accurate}. Please refer to the top-left of Fig. \ref{img:weightsRho}), while when $\rho=0.07$, only three component super-resolvers (the 4-th, 8-th, and 9-th methods, \emph{i.e.}, IA~\cite{timofte2016seven}, VDSR~\cite{kim2016accurate}, and DRCN~\cite{kim2016deeply}) contribute to the final result. (ii) \emph{The prior knowledge learned from the reference dataset is effective}. This can be concluded by comparing the results when $\rho=0.07$ and $\rho=100$. When $\rho=100$, all the component super-resolvers will be treated equally, \emph{i.e.}, the ensemble weights are set to the same value (please refer to the bottom-right of Fig. \ref{img:weightsRho} )), the performance is worse. This can be illustrated by that the poor component super-resolver with unreasonable reconstruction of the results will pull down the overall reconstruction performance.

From Fig. \ref{img:PlotLambda}, we can learn that the performance increases with the increase of the value of $\lambda$, and then slightly decrease. This indicates that the prior knowledge of the reference ensemble weights is very effective for the SR reconstruction. When $\lambda=0$, it reduces to the case of considering only the reconstruction constraint. There is 0.2 dB gain of the proposed method over the method neglecting the prior knowledge of the reference ensemble weights. The decrease after $\lambda=0.8$ is because of overemphasizing the prior knowledge of the reference ensemble weights while neglecting the reconstruction constraint. This verifies our motivation of simultaneously taking into consideration of the reconstruction constraint (favors the degenerate model) and the prior knowledge generated from the reference dataset.

For the sake of convenience comparisons, in Table \ref{tab:test} we tabulate the performance of above-mentioned cases: RefESR without reconstruction constraint, RefESR without weights prior, ensemble via averaging, and the proposed RefESR method. In the second row, we also list the performance of the best component super-resolver, \emph{i.e.}, VDSR~\cite{kim2016accurate}. The two cases of introducing the reconstruction constraint and averaging based ensemble obtain the similar results, which is consistent with Wang \emph{et al}.'s results (see the Table 4 in \cite{wang2017ensemble}). It also shows that  it is not enough to consider reconstruction constraint alone. When compared with RefESR without reconstruction constraint and RefESR without weights prior, it indicates that the ensemble weight prior is effective and relatively more important than the reconstruction constraint. This is mainly because that the component super-resolvers used in our experiments are very competitive and have very good SR performance, and these methods essentially satisfy the reconstruction constraint. {By incorporating the prior knowledge of ensemble weights, our method has a quite impressive gain, \emph{i.e.}, 0.2 dB. For image SR is a very hot topic and becomes a test bed for many emerging models and algorithms, and some very superior methods are constantly being presented, and thus it is very difficult for one new method to obtain a very large gain over previous methods.} {From Table \ref{tab:test}}, we can also see that simply averaging all the results of different methods will sacrifice the final ensemble performance, \emph{e.g.}, 0.07 dB decrease when compared with VDSR~\cite{kim2016accurate}. This once again shows the effectiveness of adaptively assigning different ensemble weights to different component super-resolvers.

Under above optimal parameter settings, $\rho=0.07$ and $\lambda=0.8$, we examine the final ensemble weights of different testing images on the \textbf{SET14}. As shown in Fig. \ref{img:weight14}, three best component super-resolvers, IA~\cite{timofte2016seven}, VDSR~\cite{kim2016accurate}, and DRCN~\cite{kim2016deeply}, dominate the SR reconstruction. The better the quality of SR performance over the reference dataset is, the larger the ensemble weight is. This verifies our assumption that better performance on the reference dataset should get a relatively larger weight when reconstructing the HR output image of an LR input one in the ensemble framework. Moreover, from the results we also can see that some component super-resolvers with low quality do not contribute substantially to the results. When we only select three methods that play dominant roles (i.e., the ensemble weights of these methods are relatively large), we find that this has little impact on the final performance of the proposed algorithm. This is consistent with the observation that it may be better to combine some instead of all of the component super-resolvers.

\subsection{Compare with State-of-the-art}
To verify the effectiveness of the proposed RefESR method, we provide quantitative and qualitative comparisons with the {eight} component super-resolvers and Wang \emph{et al.}'s ESCN method~\cite{wang2017ensemble} over \textbf{SET5}, \textbf{SET14}, {and \textbf{Urban100}} for different upscaling factors. {We add the visual results of Bicubic interpolation, which can be seen as the baseline.} {In Table \ref{tab:PSNRSSIMSet14}, Table \ref{tab:PSNRSSIMSet5}, and Table \ref{tab:PSNRSSIMUrban100}}, we show the PSNR and SSIM for adjusted anchored ten comparison methods and our RefESR method. {All the values in tables are the average over all the images within a dataset. From the results , we can learn that} our method outperforms almost all existing methods, including the most competitive deep learning based methods, in all datasets and scale factors (in term of PSNR). Only in two situations our RefESR method is just a little worse than DRCN~\cite{kim2016deeply} (in term of SSIM). {The visual comparisons of three typical images are shown in Fig. \ref{fig:zebra}, Fig. \ref{fig:ppt}, and Fig. \ref{fig:woman}. To make the comparison more notable, we also give the local region (marked by red boxes) magnification results.} Our method produces relatively shaper boundaries and is free of the ringing artifacts.This can be explained as the following two reasons: (i) Through the ensemble strategy, it is possible to highlight the good side of these approaches with superior performance while inhibiting the poor side these approaches with poor performance. (ii). The reconstruction artifacts, ringing artifacts, of one component super-resolver can be weakened by fusing multiple results. But we must also see that if all methods produce ringing artifacts in the same region, the ensemble results cannot eliminate these artifacts.

Image SR is a very hot topic and becomes a test bed for many emerging models and algorithms, especially recently very popular deep learning {techniques}. Almost every few days there will be a new algorithm is released in \emph{arXiv}. In the process of preparing this paper, a series of deep learning based SR algorithms are released and achieve very good performance. To further demonstrate the effectiveness of the proposed ensemble learning framework, we additionally ensemble the most competitive method, EDSR \cite{lim2017enhanced}, with aforementioned nine component super-resolvers. Table \ref{tab:PSNRSSIMThreeDB} shows the results of EDSR and the proposed RefESR. In addition, we also give the results of the combination of geometric ensemble strategy and the proposed ensemble strategy, which is denoted as RefE$^2$SR. From these results, we observe that: (i) Although the performance of EDSR is already very good, the proposed ensemble framework can still improve the final results. It shows that EDSR and other methods still have a certain degree of complementarity. (ii) RefE2SR is better than RefESR. This can be explained by the following reasons: when geometric ensemble strategy is applied to the component super-resolvers, their performance can be improved. With these improved SR results, our proposed method can further promote the overall performance of the combination ensemble strategy.

\begin{table}[t]
\centering
\caption{\textcolor[rgb]{0.00,0.00,1.00}{PSNR (dB) and SSIM results by ensembling EDSR and the other nine component super-resolvers}.}
\label{tab:PSNRSSIMThreeDB}
\vspace{1pt}
\scriptsize
\begin{tabular}{|c||c|c|c|c|c|c|}
\hline
Scale &        	\multicolumn{2}{c|}{$\times 2$} & \multicolumn{2}{c|}{$\times 3$} &	\multicolumn{2}{c|}{$\times 4$} \\
\hline
Metric &	PSNR&SSIM&PSNR&SSIM&PSNR&SSIM\\
\hline
\hline
Dataset      & \multicolumn{6}{c|}{\textbf{SET14}}\\
\hline
EDSR \cite{lim2017enhanced}	&33.68	&0.9172	&30.34	&0.8434	&28.66  & 0.7845\\
RefESR	&33.85	&0.9194	&30.45	&0.8454	&28.75    & 0.7862\\
RefE$^2$SR	&33.95	&0.9203	&30.61	&0.8470	&28.91 & 0.7873\\
\hline
\hline
Dataset      & \multicolumn{6}{c|}{\textbf{SET5}}\\
\hline
EDSR \cite{lim2017enhanced}	&38.11 	&0.9601 	&34.64 	&0.9282 	&32.46 	&0.8968\\
RefESR	&38.16 	&0.9607 	&34.66 	&0.9285 	&32.48 	&0.8970\\
RefE$^2$SR	&38.26 	&0.9611 	&34.92 	&0.9299 	&32.77 	&0.8996\\
\hline
\hline
Dataset      & \multicolumn{6}{c|}{\textbf{Urban100}}\\
\hline
EDSR \cite{lim2017enhanced}	&32.93 &	0.9351 &	28.80 &	0.8653 &	26.64 	&0.9033\\
RefESR	&33.02 &	0.9373 &	28.89 &	0.8672 &	26.73 	&0.9039\\
RefE$^2$SR	&33.19 	&0.9378 &	29.03 &	0.8698 &	26.96 	&0.9086\\
\hline
\end{tabular}
\end{table}

\begin{table}[t]
\centering
\footnotesize
 \caption{Average PSNR (dB) and SSIM comparisons of different methods under different noise levels.}
  \begin{tabular}{lcccccc}
  \toprule
\multirow{2}{*}{Methods}	&	\multicolumn{2}{c}{$\sigma=0$}	&	\multicolumn{2}{c}{$\sigma=5$}&	\multicolumn{2}{c}{$\sigma=10$}	\\
  \cline{2-7}
	&	PSNR	&	SSIM	&	PSNR	&	SSIM	&	PSNR	&	SSIM	\\
\midrule

Wang~\cite{Wang2005Eig}	&	27.73 	&	0.7642 	&	27.93 	&	0.7564 	&	27.01 	&	0.7251 	\\
NE~\cite{Chang2004NE}	&	30.73 	&	0.8587 	&	29.19 	&	0.8065 	&	27.92 	&	0.7682 	\\
LSR~\cite{Ma2010LSR}	&	32.12 	&	0.8969 	&	28.70 	&	0.7469 	&	24.44 	&	0.5269 	\\
SR~\cite{Yang2010TIP}	&	32.21 	&	0.8983 	&	28.37 	&	0.7238 	&	23.96 	&	0.4903 	\\
LcR~\cite{Jiang2014LcRTMM}	&	32.23 	&	0.8981 	&	30.09 	&	0.8275 	&	30.29 	&	0.8449 	\\
SSR~\cite{jiang2017noise}	&	32.34 	&	0.8992 	&	29.82 	&	0.8445 	&	28.56 	&	0.8022 	\\
DRP~\cite{shi2015kernel}	&	32.60 	&	0.9213 	&	27.79 	&	0.7102 	&	23.21 	&	0.4585 	\\
RefESR	&	33.13 	&	0.9252 	&	30.67 	&	0.8500 	&	30.98 	&	0.8624 	\\
\midrule
Gains	&	0.53 	&	0.0039 	&	0.58 	&	0.0055 	&	0.69 	&	0.0175 	\\

  \bottomrule
  \label{tab:face}
 \end{tabular}
\end{table}

\subsection{Ensemble Super-Resolution Results with Face Images}
In order to verify the universality of our proposed ensemble framework, we test our the proposed RefESR method on the task of face image SR, a.k.a. face hallucination \cite{Baker2000}. Similarly, through a reference set, the performance of different face SR algorithms is learned, i.e., their ensemble weights are estimated, and then the reconstruction results of different algorithms on the newly observed LR faces are integrated based on the estimated weights.

The reference face dataset consists of 600 images of 600 subjects, in which 200 subjects are from CAS-PEAL-R1 face database \cite{Gao2007CAS}, 100 subjects are from CUHK face database \cite{wang2009face}, 200 subjects are from COX-S2V face database \cite{huang2015benchmark}, and 100 subjects are from Scface face database \cite{grgic2011scface}. To evaluate the performance of the component super-resolvers, we additionally collect 20, 10, 20, and 10 face images from these four databases to form the evaluation dataset. For testing, we capture 42 High-Definition (HD) images, whos face images are very different from the face image in reference face dataset. Some example images are shown in Fig.~\ref{fig:HDimages}. In our experiments, the component super-resolvers for face images include Wang \emph{et al}.'s Eigentrasformation method \cite{Wang2005Eig}, neighbor embedding (NE) \cite{Chang2004NE}, least squares representation (LSR) \cite{Ma2010LSR}, sparse representation (SR) \cite{Yang2010TIP}, locality-constrained representation (LcR) \cite{Jiang2014LcRTMM}, smooth sparse representation (SSR) \cite{jiang2017noise}, dual regularization prior (DRP) \cite{shi2015kernel}.

Similarly, we apply the reference face dataset to train the component super-resolvers and use the evaluation dataset to obtain their performance in terms of PSNR and SSIM. Therefore, the the reference weight vector $\textbf{w}^{ref}$ can be calculated according to Eq. (\ref{eq:wref0}). Based on the prior knowledge of $\textbf{w}$, we can obtain the optimal ensemble weight vector for each input LR face image by \ref{eq:Function2}. In addition, we also conduct some experiments to test the robustness of our method when the input is contaminated by noise. Our first impression is: given the noise input, if the resulting images generated by different algorithms are not optimal (may contain noise), then the noise can be smoothed through fusion of different results.

Table \ref{tab:face} tabulates the performance (in terms of average PSNR and SSIM) of different component super-resolvers and the proposed RefESR method under different noise levels, i.e., $\sigma = 0, 5, 10$. We learn that RefESR achieves the best average PSNR and SSIM results. The gains of the proposed method over the second best method are obvious, greater than 0.5 dB in term of PSNR. In addition, we also observe that with the increase of noise, the advantage of the proposed method is much more obvious. In particular, when the input is noiseless, the PSNR gain of the proposed method over the second best method is 0.53 dB. When the input is contaminated by different levels of noise, the gain is 0.58 dB for $\sigma=5$ and 0.69 dB for $\sigma=10$, respectively. We attribute this to the advantages of ensemble learning, which can eliminate the uncertainties caused by noise in different methods. Fig. \ref{img:faceresults} shows some visual comparison results of component super-resolvers and the proposed method. From these results, we observe that the proposed RefESR method can remove most of the noise and well maintain the main structural information.

\begin{figure}[h]
\centering
\centerline{\includegraphics[width=8.5cm]{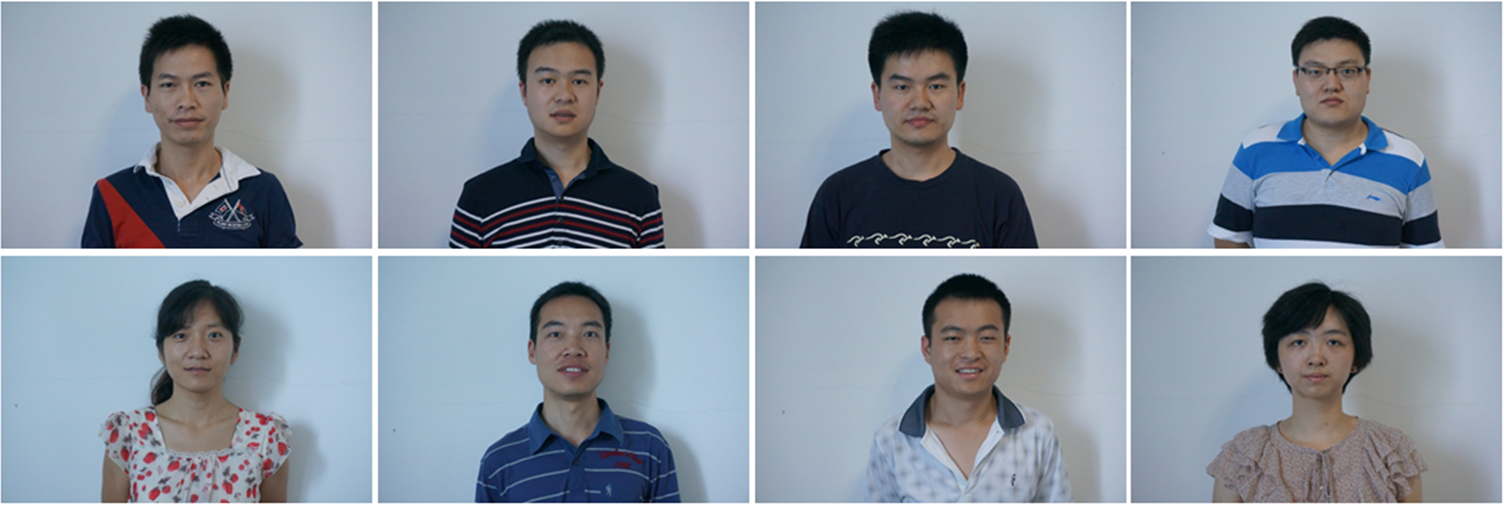}}
\vspace{-0.20cm}
\caption{Eight real-world images used to test the performance of different face hallucination methods. These images captured by an HD camera in the normal night condition. The eight face images are indexed as Img1 to Img8 in the following.}
\label{fig:HDimages}
\end{figure}

 \begin{figure*}
  \centering
  \includegraphics[width=0.9750\textwidth]{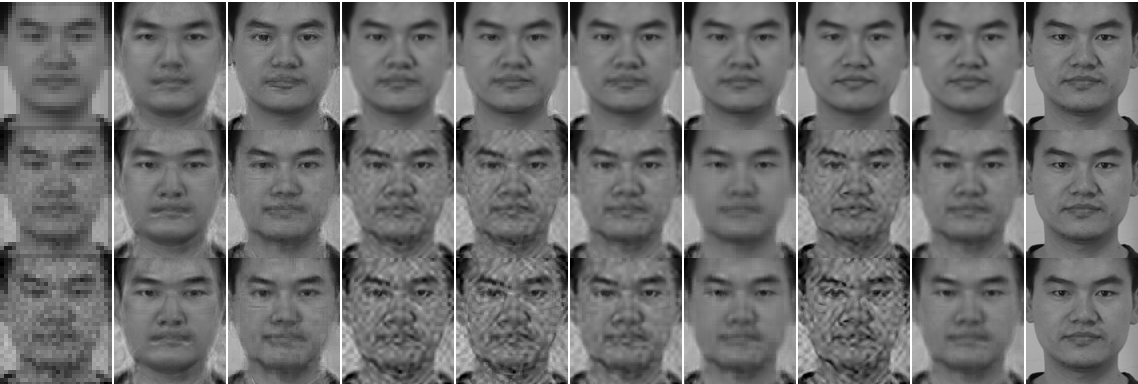}\\
  \vspace{-0.20cm}
  \caption{Face SR results of the component super-resolvers and the proposed RefESR method on one individual. From left to right and top to bottom: the LR testing faces, hallucinated results of Bicubic interpolation method, Wang \emph{et al}.'s method \cite{Wang2005Eig}, NE \cite{Chang2004NE}, LSR \cite{Ma2010LSR}, SR \cite{Yang2010TIP}, LcR \cite{Jiang2014LcRTMM}, SSR~\cite{jiang2017noise}, DRP~\cite{shi2015kernel}, and the proposed RefESR method. The last column is the HR ground truth. The first row is the results without noise, while the second and the third rows are the results when the noise level ($\sigma$) is 5 and 10, respectively.}
  \label{img:faceresults}
  \vspace{-0.20cm}
\end{figure*}

\section{Discussion}
\label{sec:dis}

In this section, we show deep analysis to the proposed ensemble learning framework, so that readers can better capture our idea.

\textbf{Time complexity}. By ensembling the results of some state-of-the-art methods, we can expect better reconstruction performance. This will also result in very high computational complexity. Despite the efficient solution of the optimization procedures of ESR, which take around 0.06 seconds for each image, the computational complexity of our method is high because the total running time is the sum of (i) all component super-resolvers and (ii) the optimization procedures of ESR. Therefore, the computational complexity will be a bottleneck for our approach in practical applications.

\textbf{Theoretical guarantee}. Another drawback of the proposed algorithm is that there is no theoretical guarantee to produce a better result by ensembling different methods, which is also the limitation of conventional ensemble learning based machine learning methods~\cite{zhou2012ensemble}. 
From the experiments, we learn that in most cases our RefESR method beats all the comparison methods. However, under some situations, our RefESR method is worse than the best comparison method. Therefore, in the future we will consider the learning of a \emph{safe prediction} from multiple component super-resolvers, which is not worse than the performance of all component super-resolvers.

\textbf{Model universality}. Different methods can adapt to different kind of test images. For example, there are SR algorithms for general images and SR algorithms for specific images such as digital characters, faces, and irises. SR models trained on general images are not suitable for reconstruction of specific images, and vice versa. Furthermore, the ensemble weight prior (of ensemble learning) obtained from the general images of different methods may not necessarily reflect the SR ability on specific images. In this paper, the proposed ensemble learning based SR method is applied to the general images and face images SR tasks. Through the experiments, we believe that the proposed method is indeed effective in the sense of improving the performance of the existing generic image SR algorithms or face image SR algorithms. In summary, the proposed framework is very universal in the sense that given a reference dataset, the proposed method can improve the performance of existing SR methods when the input image is with the same class of the reference dataset.

\textbf{Choice of component super-resolvers}. In this paper, we do not consider the complementarity of different methods, but directly select several representative methods in the current SR field, including four shadow learning-based methods and five deep learning-based methods. We also believe that when choosing component super-resolvers, it should consider the characteristics of different algorithms. Ensembling component super-resolvers with different characteristics is more likely to improve the final ensemble performance.

\textbf{Global reconstruction constraint}. As shown in many previous works~\cite{yang2012single, Zeyde2012,Yang2010TIP}, global reconstruction constraint, which claims that the degenerated HR estimation should be consistent with the observed LR image~\cite{Yang2010TIP, Gao2012Joint}, is very effective for enhancing the final super-resolved results by an iterative back projection strategy. In our experiments, we have found that if the performance of the component super-resolver is good enough, the improvement brought by reconstruction constraint is very limited. In other words, when the component super-resolver is good enough, it can basically meet the reconstruction constraint.

\section{Conclusion}
\label{sec:conclusion}
In this paper, we present a novel framework based on ensemble learning to solve the single image SR problem. It introduces a reference dataset and incorporates the learned prior of each component super-resolver, which states that the method obtains a better performance on the reference dataset should get a relatively larger weight when reconstructing the HR output image of an LR input one in the ensemble framework, to regularize the optimization of ensemble weights. We simultaneously model this learned prior of ensemble weights and reconstruction constraint, which states that the degenerated HR image should be equal to the LR observation one, by an MAP formulation. Finally, we present an analytical solution to this constrained least squares problem induced from the MAP framework. Results show the effectiveness of the introduced prior knowledge of ensemble weights learned from a reference dataset.

\section*{Acknowledgment}
We would like to thank Dr. Zehao Huang, the author of \cite{wang2017ensemble}, for his kind providing of the results of ESCN algorithm. We also would like to thank Dr. Jingang Shi, the author of \cite{shi2015kernel}, for his kind providing of the source codes of DRP based face super-resolution approach.

\bibliographystyle{IEEEtran}
\bibliography{RefESR}

\end{document}